\documentclass[lettersize,journal]{IEEEtran}

\usepackage{verbatim}
\usepackage{soul}
\usepackage{cite}
\usepackage{threeparttable}
\usepackage{amsmath,amssymb,amsfonts}
\usepackage{tabularx}
\usepackage{algorithm}             
\usepackage{algorithmic}
\usepackage{graphicx}
\usepackage{textcomp}
\usepackage{xcolor}
\usepackage[algo2e]{algorithm2e}
\def\BibTeX{{\rm B\kern-.05em{\sc i\kern-.025em b}\kern-.08em
    T\kern-.1667em\lower.7ex\hbox{E}\kern-.125emX}}
\newtheorem{theorem}{Theorem}[section]

\newtheorem{lemma}{Lemma}[section]
\newtheorem{definition}{Definition}[section]
\newenvironment{proof}{{ \noindent\it Proof :}\quad}{\hfill $\square$\par}
\usepackage{subfigure}

\begin{document}

%
\title{Multi-class Classification with Fuzzy-feature Observations: Theory and Algorithms}
%
%
%

\author{Guangzhi~Ma,
        Jie~Lu,~\IEEEmembership{Fellow,~IEEE,} 
       Feng~Liu,~\IEEEmembership{Member,~IEEE,} \\
       Zhen~Fang,~\IEEEmembership{Member,~IEEE,}~and~Guangquan~Zhang
\thanks{The work presented in this paper was supported by the Australian Research Council (ARC) under FL190100149.

Guangzhi Ma, Jie Lu, Feng Liu, Zhen Fang and Guangquan Zhang are with Australian Artificial
Intelligence Institute, Faulty of Engineering and Information Technology,
University of Technology Sydney, Sydney, NSW, 2007, Australia, e-mail: Guangzhi.Ma@student.uts.edu.au,
\{Jie.Lu; Feng.Liu; Zhen.Fang; Guangquan.Zhang\}@uts.edu.au.}
}

%
%

\markboth{IEEE TRANSACTIONS ON CYBERNETICS,~Vol.~XX, No.~XX, XX~2022}%
{Ma \MakeLowercase{\textit{et al.}}: Multi-class Classification with Fuzzy-feature Observations: Theory and Algorithms}


\maketitle

\begin{abstract}
The theoretical analysis of multi-class classification has proved that the existing multi-class classification methods can train a classifier with high classification accuracy on the test set, when the instances are \emph{precise} in the training and test sets with same distribution and enough instances can be collected in the training set. However, one limitation with multi-class classification has not been solved: how to improve the classification accuracy of multi-class classification problems when only imprecise observations are available. Hence, in this paper, we propose a novel framework to address a new realistic problem called \emph{multi-class classification with imprecise observations} (MCIMO), where we need to train a classifier with fuzzy-feature observations. Firstly, we give the theoretical analysis of the MCIMO problem based on fuzzy Rademacher complexity. Then, two practical algorithms based on support vector machine and neural networks are constructed to solve the proposed new problem. Experiments on both synthetic and real-world datasets verify the rationality of our theoretical analysis and the efficacy of the proposed algorithms. 
\end{abstract}


%
\IEEEpeerreviewmaketitle

\section{Introduction}
%
%
%
%
\IEEEPARstart{M}{achine} learning methods for the multi-class classification problem have gained great achievements in many areas, including medical imaging \cite{2019seebock}, natural language processing \cite{2013xia}, biology \cite{2016zhu} and computer vision \cite{nanni2017handcrafted}. The theoretical analysis of existing well-known multi-class classification machine learning algorithms, such as \emph{support vector machine} (SVM) \cite{wang2018transfer} and neural networks \cite{chen2013fuzzy}, has been well researched \cite{Mehryar2012}. Recently, many researchers considered using different measures to give the estimation error bounds for classification problems that can guarantee the rationality of these algorithms. These measures include Rademacher complexity \cite{Panchenko2002Empirical,Mehryar2012,Maximov2018}, VC-dimension \cite{Allwein2000Reducing,2014Optimal}, stability and \emph{probably approximately correct} (PAC)-Bayesian \cite{hardt2016train,mcallester2013pac}, and local Rademacher Complexity \cite{Xu2016,Li2018}.

Rademacher complexity is a crucial tool to derive generalization bounds, which measure how well a given hypothesis set can fit random noise. A Rademacher complexity based bound was first proposed by Koltchinskii and Panchenko \cite{Panchenko2002Empirical}. Subsequently, this bound was improved in \cite{Mehryar2012}. Then, Maximov, Amini and Harchaoui \cite{Maximov2018} presented a new estimation error bound using  Rademacher complexity for multi-class classification issues. In addition, to ensure multi-class PAC learnability, a series of estimation error bounds based on VC-dimension and Natarajan dimension were proposed in \cite{Allwein2000Reducing,2014Optimal}. Because of the dependence on dimensions, these VC-dimension based bounds rarely apply to large-scale issues. To conduct theoretical analysis of neural networks for multi-class classification problems, Hardt \emph{et al.} \cite{hardt2016train} and McAllester \cite{mcallester2013pac} introduced the new bounds based on stability and PAC-Bayesian. Further, tighter and sharper bounds were proposed in \cite{Xu2016,Li2018} by using local Rademacher complexity. According to these theoretical analyses, it illustrates that we can always learn a good classifier for multi-class classification problems to predict the test set when the instances are precise in the training and test sets with same distribution and enough instances can be collected in the training set.

However, there is one limitation with multi-class classification that the existing methods can not handle the scenario that only imprecise observations are available. For example, the readings on many measuring devices are not exact numbers but intervals because there are only a limited number of decimals available on most of these measuring devices. Thus, this scenario has inspired us to consider a further realistic problem called \emph{multi-class classification with imprecise observations} (MCIMO). With the MCIMO problem, we aim to train a classifier with high classification performance for multi-class classification problems when the features of all the instances in both training and test sets are imprecise (e.g., fuzzy-valued or interval-valued features). 

The main challenge to solving the MCIMO problem is how to handle observations with fuzzy-valued or interval-valued features. Existing well-known machine learning methods can not be directly used to address the MCIMO problem. Recently, combining fuzzy techniques with machine learning methods (especially for transfer learning methods \cite{zhang2020clarinet,fang2020open,zhong2021does,zhen2022heterogeneous,jiahua2022domain}) has drawn increasing attention. In the literature review section, we will give a brief review of these machine learning methods with fuzzy techniques \cite{Colubi2011Nonparametric, wang2020deep, liu2020multi, zuo2019fuzzy2, lu2019fuzzy, ma2021learning}. According to these fuzzy-based methods, it demonstrates that fuzzy techniques are powerful tools to analyze imprecise observations and provide better interpretability to handle the uncertainty of different issues. Therefore, we consider using fuzzy techniques to address the MCIMO problem because they can represent the imprecise features of the instances in both training and test sets and can handle different types of uncertainty issues.

In this paper, we consider using fuzzy random variable, which was proposed in \cite{Ralescu1986,Wu1999Probability}, to represent the imprecise feature of the instances. Then, we give the theoretical analysis and obtain the estimation error bounds for the MCIMO problem. In the MCIMO problem, these bounds are really important as it ensures that we can always train a fuzzy classifier with high classification accuracy when the instances are drawn from the same fuzzy distribution and enough fuzzy-feature instances can be collected. 

Subsequently, we construct two fuzzy technique-based algorithms, which combine fuzzy techniques with SVM and neural networks to analyze fuzzy data. The proposed algorithms contain two main parts. The first part aims to extract the most significant crisp-valued information from imprecise observations, which is the main difficulty of the proposed algorithms. In this paper, we compare the performance of different defuzzification methods on synthetic datasets to find the optimal defuzzification function for the proposed algorithms. The second part is to classify the extracted crisp-valued information by two well-known machine learning methods: SVM and neural networks. In addition, interval-valued data is also a common type of imprecise data in real-world scenarios. In this paper, we give one approach to apply the proposed methods to analyze interval-valued data. Finally, experimental results on both synthetic and real-world datasets reveal the superiority of the proposed algorithms and demonstrate that the proposed fuzzy-based methods can obtain better performance to analyze fuzzy data or interval-valued data than non-fuzzy methods through comparisons with seven baselines. The main contributions of this paper are as follows.
\begin{enumerate}
\item We identify a novel problem called MCIMO, which considers addressing the multi-class classification problem when only imprecise observations are available, and we propose a framework to handle this problem. Based on this framework, two fuzzy technique-based machine learning algorithms called DF-SVM and DF-MLP are constructed, which combine fuzzy techniques with SVM and neural networks. These algorithms significantly improve classification accuracy since they use fuzzy vectors to express the distribution of imprecise data and apply different defuzzification methods to extract crisp-valued information from imprecise observations.
\item We give the theoretical analysis of the MCIMO problem based on the fuzzy Rademacher complexity, which ensures that we can always train a fuzzy classifier with high classification accuracy. This theory provides a theoretical basis for fuzzy data analysis.
\item By comparing the performance of different defuzzification methods on synthetic datasets, we find the optimal defuzzification function for the fuzzy technique-based SVM and neural networks algorithms. Through experimental comparisons with several baselines on both synthetic and real-world datasets, it demonstrates the superiority of the proposed algorithms to analysis fuzzy data and interval-valued data. 

\end{enumerate}

The remainder of this paper is structured as follows. Section \ref{sec:Literature Review} presents a brief review of the methods which combine fuzzy techniques with machine learning methods. Section \ref{sec:preliminaries} introduces the related definitions. Section \ref{sec:MCIMO} introduces and gives a formal definition of the MCIMO problem. Section \ref{sec:Theoretical Analysis} gives the theoretical analysis of the MCIMO problem. Section \ref{sec:algorithm} proposes a novel framework to address the MCIMO problem and constructs two algorithms based on this framework to analyze fuzzy-feature observations. In Sections \ref{sec:experiment} and \ref{sec:experiments}, the experiments on both synthetic and real-world datasets are constructed to show the superiority of the proposed algorithms. Section \ref{sec:conclusion} concludes this paper and outlines future work.

\section{Literature Review}
\label{sec:Literature Review}
In this section, a brief review of the methods which combine fuzzy techniques with machine learning methods is presented.

On the one hand, for classification tasks, Colubi \emph{et al.} \cite{Colubi2011Nonparametric} integrated fuzzy $L_2$ metrics \cite{Sinova2014A} with the discriminant analysis approach to analyze fuzzy data. Yang \emph{et al.} \cite{Yang2011fuzzysvm} proposed a novel fuzzy SVM algorithm based on a kernel fuzzy c-means clustering method to deal with the classification problems with outliers or noises. Rong \emph{et al.} \cite{Rong2007Classification} introduced a new classification method, which applies the defuzzified Choquet integral to address heterogeneous fuzzy data classification issues.  Wang \emph{et al.} \cite{wang2020deep} presented a novel deep-ensemble-level-based \emph{Takagi–Sugeno–Kang} (TSK) fuzzy classifier to address imbalanced data classification tasks, which achieved both promising classification performance and high interpretability of zero-order TSK fuzzy classifiers. Liu \emph{et al.} \cite{liu2020anovel} used fuzzy vectors to model imprecise observations of distributions and help address the two-sample testing problem that is a core problem in the machine learning field \cite{gretton2012kernel,liu2020learning,liu2021meta}.

In addition, in the area of transfer learning, Behbood \emph{et al.} \cite{behbood2014fuzzy,behbood2015multistep} proposed a series of novel fuzzy-based transfer learning methods for long-term bank failure prediction, which use the fuzzy sets and the concepts of similarity and dissimilarity to modify the labels of the target instances. Deng \emph{et al.} \cite{yang2015takagi,deng2018transductive,xie2018generalized,jiang2021eeg} proposed 
several new approaches that integrate \emph{TSK fuzzy system} (TSK-FS) with transfer learning to recognize epileptic electroencephalogram signals. To solve the \emph{heterogeneous unsupervised domain adaptation} (HeUDA) problems for classification tasks, Liu \emph{et al.} \cite{liu2018unsupervised} introduced a novel HeUDA approach utilizing shared fuzzy equivalence relations via fuzzy geometry, which can measure the similarity between the features of the instances in the source and target domain. Further, \cite{liu2020multi} enhanced this method, which called shared-fuzzy-equivalence-relations neural network, to analyze another challenging problem called the multi-source heterogeneous unsupervised domain adaptation. 

In contrast, for regression tasks Deng \emph{et al.} \cite{deng2012knowledge,deng2013knowledge} proposed several novel transfer learning approaches utilizing the Mamdani-Larsen fuzzy systems and TSK-FS. Further, the authors \cite{deng2016enhanced} improve the above model to construct a new transfer learning model that uses two knowledge-leverage strategies, learning from the TSK-FS model, to enhance the two types of parameters for the target domain. In addition, Zuo \emph{et al.} \cite{zuo2018granular} applied granular computing techniques to transfer learning and proposed a comprehensive domain adaptation framework based on the T–S fuzzy model. Subsequently, \cite{zuo2019fuzzy2} presented a novel fuzzy rule-based transfer learning model, which integrates an infinite Gaussian mixture model with active learning. Applying these two techniques, researchers can identify the data structure and select an appropriate source domain when multi-source domains are available, and choose labeled data for the target model with high efficiency when the target domain contains insufficient data. Hence, Lu \emph{et al.} \cite{lu2019fuzzy} presented a novel fuzzy rule-based transfer learning approach that merges fuzzy rules from multi-source domains in both homogeneous and heterogeneous scenarios. Besides, some new fuzzy-based clustering methods were presented in \cite{Benjamin2016Clustering,Pierpaolo2020Fuzzy} to analyze fuzzy data.

In our previous work \cite{ma2021learning}, we proposed one algorithm to solve a novel classification problem that the instances in training and test sets are all imprecise and we give the theoretical analysis of this problem. However, there are two drawbacks in our previous works. First, one gap has not be solved that there is no research to explore properties of different defuzzification methods. Second, we only verified the performance of the proposed algorithm on the synthetic dataset, while the performance of the proposed algorithm on real-world datasets is indispensable. In this paper, we address both drawbacks in our previous work.

\section{Preliminary}
\label{sec:preliminaries}
In this section, some related definitions are introduced, including the definitions of fuzzy probability density function and fuzzy probability distribution.

\begin{definition}[\cite{Wu1999Probability}]
Let $R$ be the universal set, $\widetilde{X}$ is a fuzzy random variable. Suppose $f_{\widetilde{X}_{\alpha}}(x)$ is the probability density function of $\widetilde{X}_{\alpha}^{L}$ and $\widetilde{X}_{\alpha}^{U}$, where $[\widetilde{X}_{\alpha}^{L},\widetilde{X}_{\alpha}^{U}]$ is the $\alpha$-cut of $\widetilde{X}$. We define $\widetilde{f}(\widetilde{x})$ as the fuzzy probability density function of $\widetilde{X}$. Then, the membership function of $\widetilde{f}(\widetilde{x})$ is defined as:
\begin{equation}
\begin{array}{clc}
\mu_{\widetilde{f}(\widetilde{x})}(r)=\sup\limits_{0\leq\alpha\leq1}{\alpha1_{A_{\alpha}}(r)}.
\end{array}
\end{equation}
where
\begin{equation*}
\begin{array}{clc}
A_{\alpha}& =&[\min\limits_{x\in[\widetilde{x}_{\alpha}^{L},\widetilde{x}_{\alpha}^{U}]}{f_{\widetilde{X}_{\alpha}}(x)},\max\limits_{x\in[\widetilde{x}_{\alpha}^{L},\widetilde{x}_{\alpha}^{U}]}{f_{\widetilde{X}_{\alpha}}}(x)]\\
 
  &= &[\min\{\min\limits_{\alpha\leq\beta\leq1}{f_{\widetilde{X}_{\alpha}}(\widetilde{x}_{\beta}^{L})},\min\limits_{\alpha\leq\beta\leq1}{f_{\widetilde{X}_{\alpha}}(\widetilde{x}_{\beta}^{U})}\},\\
 
 & & \max\{\max\limits_{\alpha\leq\beta\leq1}{f_{\widetilde{X}_{\alpha}}(\widetilde{x}_{\beta}^{L})},\max\limits_{\alpha\leq\beta\leq1}{f_{\widetilde{X}_{\alpha}}(\widetilde{x}_{\beta}^{U})}\}],
\end{array}
\end{equation*}

\end{definition}

\begin{definition}[\cite{ma2021learning}]
We denote $\widetilde{D}$ as the fuzzy probability distribution of $\widetilde{X}\in\mathcal{F}_{\mathbb{R}}$ (denoted as $\widetilde{X}\sim\widetilde{D}$), which contains the value range and fuzzy probability density function of $\widetilde{X}$, where $D$ represents the value range of real-valued variable $x$ which induce all fuzzy real numbers in $\widetilde{D}$. 
\end{definition}

Let $\widetilde{X}=(\widetilde{x}_{1},\widetilde{x}_{2},\cdots,\widetilde{x}_{p})\in \mathcal{F}^{p}_{\mathbb{R}^{p}}$ be $p$-fuzzy random vector, where $\widetilde{x}_{1},\widetilde{x}_{2},\cdots,\widetilde{x}_{p}\in \mathcal{F}_{\mathbb{R}}$ are i.i.d fuzzy random variables. Suppose the probability density function of $\widetilde{x}_{j}$ is $\widetilde{f}_{j}(\widetilde{x})$, $j=1,\cdots,p$. We denote the joint probability density function of $\widetilde{X}$ is $\widetilde{f}_{\widetilde{X}}(\widetilde{x})=\widetilde{f_{1}}(\widetilde{x_{1}})\bigotimes\cdots\bigotimes\widetilde{f}_{p}(\widetilde{x}_{p})$ and its membership function is defined by
\begin{equation}
\begin{array}{clc}
\xi_{\widetilde{f}_{\widetilde{X}}(\widetilde{x})}(r)=\sup\limits_{0\leq\alpha\leq1}{1_{[\widetilde{f}_{\widetilde{X}}(\widetilde{x})]_{\alpha}}(r)},
\end{array}
\end{equation}
where
\begin{equation*}
\begin{array}{clc}
&[\widetilde{f}_{\widetilde{X}}(\widetilde{x})]_{\alpha}\\
=&[\prod\limits_{j=1}^{p}\min\limits_{x_{j}\in[\widetilde{x_{j}}_{\alpha}^{L},\widetilde{x_{j}}_{\alpha}^{U}]}{f_{\widetilde{x_{j}}_{\alpha}}(x_{j})},\prod\limits_{j=1}^{p}\max\limits_{x_{j}\in[\widetilde{x_{j}}_{\alpha}^{L},\widetilde{x_{j}}_{\alpha}^{U}]}{f_{\widetilde{x_{j}}_{\alpha}}}(x_{j})]\\
    
=&[\prod\limits_{j=1}^{p}\min\{\min\limits_{\alpha\leq\beta\leq1}{f_{\widetilde{x_{j}}_{\alpha}}(\widetilde{x_{j}}_{\beta}^{L})},\min\limits_{\alpha\leq\beta\leq1}{f_{\widetilde{x_{j}}_{\alpha}}(\widetilde{x_{j}}_{\beta}^{U})}\},\\
    
&\prod\limits_{j=1}^{p}\max\{\max\limits_{\alpha\leq\beta\leq1}{f_{\widetilde{x_{j}}_{\alpha}}(\widetilde{x_{j}}_{\beta}^{L})},\max\limits_{\alpha\leq\beta\leq1}{f_{\widetilde{x_{j}}_{\alpha}}(\widetilde{x_{j}}_{\beta}^{U})}\}].
\end{array}
\end{equation*}

Then, we denote $\widetilde{\mathcal{D}}$ as the fuzzy distribution over $\widetilde{\mathcal{X}} \subset \mathcal{F}^{p}_{\mathbb{R}^{p}}$, where $\widetilde{\mathcal{D}}$ contains the value range and the joint probability density function of any fuzzy vector belongs to $\widetilde{\mathcal{X}}$.

\section{Multi-class Classification with Imprecise Observations}
\label{sec:MCIMO}
In this section, we introduce the MCIMO problem. Let $\widetilde{\mathcal{X}}\subset\mathcal{F}^{p}_{\mathbb{R}^{p}}$ be the input space and $\mathcal{Y}=[1,K]$ be the output space, and let $\widetilde{\mathcal{D}}$ be an unknown fuzzy distribution over $\widetilde{\mathcal{X}}$. Suppose $\widetilde{S} = \{(\widetilde{X}_{i},y_{i})\}_{i=1}^m$ be a sample drawn from $\widetilde{\mathcal{X}}\times \mathcal{Y}$, where $\widetilde{X}_{i}=(\widetilde{x}_{i1},\widetilde{x}_{i2},\cdots,\widetilde{x}_{ip}), i=1,2,\cdots,m$ drawn i.i.d. from $\widetilde{\mathcal{D}}$ and $y_i=f(\widetilde{X}_{i})$ is the ground truth function denoted as,

\begin{equation*}
\begin{array}{clc}
    f:&\widetilde{\mathcal{X}}\rightarrow \mathcal{Y}\\
    & (\widetilde{x}_{i1},\widetilde{x}_{i2},\cdots,\widetilde{x}_{ip})\rightarrow k.
\end{array}
\end{equation*}
We noticed that if $\widetilde{X}_i\in\mathcal{X}$ belongs to the $k$th class, then $f(\widetilde{X}_i)=k$ . Let $\mathcal{H}\subset \{ h: \widetilde{\mathcal{X}}\rightarrow\ \mathbb{R}^{K}\}$ be the hypothesis set of the MCIMO problem and $\forall h\in\mathcal{H}$, 
\begin{equation*}
\begin{array}{clc}
    h:& \widetilde{\mathcal{X}}\rightarrow\ \mathbb{R}^{K}\\
    
    & (\widetilde{x}_{i1},\cdots,\widetilde{x}_{ip})\rightarrow (h_1(\widetilde{X}_i),\cdots,h_K(\widetilde{X}_i)),
\end{array}
\end{equation*}
where each $h_k(\widetilde{X}_i),k=1,\cdots,K$ represents the probability of the instance $\widetilde{X}_i$ belongs to the $k$-th category. Then, we give the definition of the loss function with respect to $h$,
\begin{equation*}
\begin{array}{clc}
    l:\mathbb{R}^{K}\times \mathcal{Y}\rightarrow\mathbb{R}_{+}.
\end{array}
\end{equation*}

Let $L_{\mathcal{H}} = \{l(h(\widetilde{X}), y) |\widetilde{X}\in \widetilde{\mathcal{X}}, h\in \mathcal{H}, y\in\mathcal{Y}\}$ be the class of loss functions associated with $\mathcal{H}$.

The traditional multi-class classification problems aim to use the sample $\widetilde{S}$ to find a hypothesis $h\in \mathcal{H}$ which can cause as small as possible risk $R(h)$ with respect to $f$. In the MCIMO problem, the purpose is similar to traditional multi-class classification problems. Then, we give the definition of the risk with respect to $h$,
\begin{equation}
\begin{array}{clc}
    R_{\widetilde{\mathcal{D}}}(h)\triangleq R(l(h(\widetilde{X}),y))=E_{\widetilde{X}\sim\widetilde{\mathcal{D}}}[l(h(\widetilde{X}),y)],
\end{array}
\end{equation}
where the notion of $E_{\widetilde{X}\sim\widetilde{\mathcal{D}}}[l(h(\widetilde{X}),y)]$ can be fund in \cite{ma2021learning}.

Thus, to address the MCIMO problem, we are committed to find the optimal hypothesis function $h^{\ast}$ to minimize the risk, i.e., $h^{\ast} = \arg\min_{h\in \mathcal{H}}{R_{\widetilde{\mathcal{D}}}(h)}$. 

\section{Theoretical Analysis of the MCIMO problem}
\label{sec:Theoretical Analysis}
In this section, the theoretical analysis of the MCIMO problem is presented. Firstly, the notion of fuzzy Rademacher complexity is introduced. Then, we obtain the estimation error bounds of the MCIMO problem, which guarantees that we can always obtain a fuzzy classifier with high classification accuracy when infinite fuzzy-feature instances are available.

\begin{definition}[\cite{ma2021learning}] Let $L_{\mathcal{H}}$ be a family of loss functions and $\widetilde{S} = \{(\widetilde{X}_{i},y_{i})\}_{i=1}^m$ a sample drawn from $\mathcal{F}^{p}_{\mathbb{R}^{p}}\times \mathcal{Y}$. Then, the empirical fuzzy Rademacher complexity of $L_{\mathcal{H}}$ and $\mathcal{H}$ with respect to the sample $\widetilde{S}$ and $\widetilde{S}_{X}=\{\widetilde{X}_{i}\}_{i=1}^m$ are defined as:
\begin{equation}
\begin{array}{clc}
    \widehat{\mathcal{R}}_{\widetilde{S}}(L_{\mathcal{H}})=E_{\vec{\sigma}}[\sup\limits_{l\in L_{\mathcal{H}}}\frac{1}{m}\sum\limits_{i=1}^{m}{\sigma_{i}l(h(\widetilde{X}_{i}),y_{i})}],\\
    \widehat{\mathcal{R}}_{\widetilde{S}_{X}}(\mathcal{H})=E_{\vec{\sigma}}[\sup\limits_{h\in \mathcal{H}}\frac{1}{m}\sum\limits_{i=1}^{m}\sum\limits_{k=1}^{K}{\sigma_{ik}h_k(\widetilde{X}_{i})}],
\end{array}
\end{equation}
 where $\vec{\sigma}=(\sigma_{1},\cdots,\sigma_{m})^{T}$, with $\sigma_{i}$s independent random variables drawn from the Rademacher distribution, i.e. $Pr(\sigma_{i} = +1)=Pr(\sigma_{i} = -1)=\frac{1}{2}, i=1,\cdots,m$.
\end{definition}

\begin{definition}[\cite{ma2021learning}]
Let $\widetilde{\mathcal{D}}^{'}\triangleq\widetilde{\mathcal{D}}\times \mathcal{Y}$ and $\widetilde{\mathcal{D}}$ denote the fuzzy distribution according to $\widetilde{S}$ and $\widetilde{S}_{X}$. Then, the fuzzy Rademacher complexity of $L_{\mathcal{H}}$ and $\mathcal{H}$ are defined as follow:
\begin{equation}
\begin{array}{clc}
    \widetilde{\mathcal{R}}_{\widetilde{S}\sim\widetilde{\mathcal{D}}^{'}}(L_{\mathcal{H}})=E_{\widetilde{\mathcal{D}}^{'}}[\widehat{\mathcal{R}}_{\widetilde{S}}(L_{\mathcal{H}})],\\
    \widetilde{\mathcal{R}}_{\widetilde{S}_{X}\sim\widetilde{\mathcal{D}}}(\mathcal{H})=E_{\widetilde{\mathcal{D}}}[\widehat{\mathcal{R}}_{\widetilde{S}_{X}}(\mathcal{H})].\\
\end{array}
\end{equation}
\end{definition}

Using related lemmas and theorems (shown in \cite{ma2021learning}) and the theoretical analysis of traditional multi-class classification algorithms (show in \cite{Mehryar2012,Allwein2000Reducing,Panchenko2002Empirical,Maximov2018,Li2018}), the estimation error bounds with hypotheses $\mathcal{H}\}$ are show in the following theorem.
\begin{theorem}[\cite{ma2021learning}]\label{T-1}
Let $\widetilde{S} = \{(\widetilde{X}_{i},y_{i})\}_{i=1}^m$ and $\widetilde{S}_{X} = \{\widetilde{X}_{i}\}_{i=1}^m, \widetilde{X}_{i}\sim\widetilde{\mathcal{D}}\in\mathcal{\widetilde{X}}, y_{i}=f(\widetilde{X}_{i})$ , and suppose that there are $C_l, C_h > 0$ such that $\sup_{h\in \mathcal{H}}{\parallel h\parallel_{\infty}}\leq C_h$ and $\sup_{\parallel h\parallel_{\infty}\leq C_h}\max_{y}{l(t,y)}\leq C_l$, and $\forall l\in L_{\mathcal{H}}$ is $L_l$-Lipschitz functions. For any $\delta>0$, with fuzzy probability at least $1-\delta$, each of the following holds for all $l\in L_{\mathcal{H}}$:
\begin{equation}
\begin{array}{clc}
    &|E_{\widetilde{X}\sim\widetilde{\mathcal{D}}}[l(h(\widetilde{X}),y)]-\frac{1}{m}\sum\limits_{i=1}^{m}l(h(\widetilde{X}_{i}),y_{i})|\\
    \leq&2\widetilde{\mathcal{R}}_{\widetilde{S}}(L_{\mathcal{H}})+C_l\sqrt{\frac{2\log(1/\delta)}{m}}\\
    &|E_{\widetilde{X}\sim\widetilde{\mathcal{D}}}[l(h(\widetilde{X}),y)]-\frac{1}{m}\sum\limits_{i=1}^{m}l(h(\widetilde{X}_{i}),y_{i})|\\
    \leq&2\widehat{\mathcal{R}}_{\widetilde{S}}(L_{\mathcal{H}})+3C_l\sqrt{\frac{2\log(2/\delta)}{m}}.
\end{array}
\end{equation}

Because $\forall l\in L_{\mathcal{H}}$ is $L_l$-Lipschitz functions, we have
\begin{equation}\label{3.3}
\begin{array}{clc}
    \widehat{\mathcal{R}}_{\widetilde{S}}(L_{\mathcal{H}})\leq\sqrt{2}L_l\widehat{\mathcal{R}}_{\widetilde{S}_{X}}(\mathcal{H})\\
    \widetilde{\mathcal{R}}_{\widetilde{S}}(L_{\mathcal{H}})\leq\sqrt{2}L_l\widetilde{\mathcal{R}}_{\widetilde{S}_{X}}(\mathcal{H}).
\end{array}
\end{equation}

Then,
\begin{equation}\label{11}
\begin{array}{clc}
    |R_{\widetilde{\mathcal{D}}}(h)-\widehat{R}_{\widetilde{\mathcal{D}}}(h)|\leq2\sqrt{2}L_l\widetilde{\mathcal{R}}_{\widetilde{S}_{X}}(\mathcal{H})+C_l\sqrt{\frac{2\log(1/\delta)}{m}}\\
    |R_{\widetilde{\mathcal{D}}}(h)-\widehat{R}_{\widetilde{\mathcal{D}}}(h)|\leq2\sqrt{2}L_l\widehat{\mathcal{R}}_{\widetilde{S}_{X}}(\mathcal{H})+3C_l\sqrt{\frac{2\log(2/\delta)}{m}}.\\
\end{array}
\end{equation}
\end{theorem}

The detailed proof of theorem \ref{T-1} can be found in \cite{ma2021learning}.

In Section \ref{sec:algorithm}, we decompose the hypothesis function into defuzzification function and optimization function. We let the loss function $l(h(\widetilde{X}_{i}), y_{i}) = l(g(M(\widetilde{X}_{i})), y_{i})$, where $g$ is a optimization function that maps $\mathbb{R}^{p}$ into $\mathbb{R}^{K}$. Let $\mathcal{M} \subset \{M:\widetilde{\mathcal{X}}\rightarrow\ \mathbb{R}^{p}\}$ be the class of defuzzification functions, $\mathcal{G}_{\mathcal{M}} \subset \{g(M(\widetilde{X})):\mathbb{R}^{p}\rightarrow\ \mathbb{R}^{K}| M\in \mathcal{M}, y\in\mathcal{Y}\}$ be the class of optimization functions associated with $\mathcal{M}$, and $L_{\mathcal{G}} = \{l(g(M(\widetilde{X}_{i})), y) |M\in \mathcal{M}, g\in \mathcal{G}, y\in\mathcal{Y}\}$ be the class of loss functions associated with $\mathcal{G}$. Then, we have:
\begin{equation}
\begin{array}{clc}
    \widehat{\mathcal{R}}_{\widetilde{S}}(L_{\mathcal{G}})=E_{\vec{\sigma}}[\sup\limits_{l\in L_{\mathcal{G}}}\frac{1}{m}\sum\limits_{i=1}^{m}{\sigma_{i}l(g(M(\widetilde{X}_{i})),y_{i})}],\\
    \widehat{\mathcal{R}}_{\widetilde{S}_{X}}(\mathcal{G}_{\mathcal{M}})=E_{\vec{\sigma}}[\sup\limits_{g\in \mathcal{G}}\frac{1}{m}\sum\limits_{i=1}^{m}\sum\limits_{k=1}^{K}{\sigma_{ik}g_k(M(\widetilde{X}_{i}))}],\\
    \widehat{\mathcal{R}}_{\widetilde{S}_{X}}(\mathcal{M})=E_{\vec{\sigma}}[\sup\limits_{M\in \mathcal{M}}\frac{1}{m}\sum\limits_{i=1}^{m}\sum\limits_{k=1}^{K}\sum\limits_{j=1}^{p}{\sigma_{ikj}M(\widetilde{x}_{ij})}]
\end{array}
\end{equation}

Then, we can get the following theorem using theorem \ref{T-1}.

\begin{theorem}[\cite{ma2021learning}]\label{T-2}
Let $\widetilde{S} = \{(\widetilde{X}_{i},y_{i})\}_{i=1}^m$ and $\widetilde{S}_{X} = \{\widetilde{X}_{i}\}_{i=1}^m, \widetilde{X}_{i}\sim\widetilde{\mathcal{D}}\in\mathcal{\widetilde{X}}, y_{i}=f(\widetilde{X}_{i})$ , and suppose that there are $C, C_l > 0$ such that $\sup_{g\in \mathcal{G}}{\parallel g\parallel_{\infty}}\leq C$ and $\sup_{\parallel g\parallel_{\infty}\leq C}\max_{y}{l(t,y)}\leq C_l$, and $\forall l\in L_{\mathcal{G}}$ is $L_l$-Lipschitz functions. For any $\delta>0$, with fuzzy probability at least $1-\delta$, each of the following holds for all $g\in L_{\mathcal{G}}$:
\begin{equation}
\begin{array}{clc}
    &|E_{\widetilde{X}\sim\widetilde{\mathcal{D}}}[l(g(M(\widetilde{X})),y)]-\frac{1}{m}\sum\limits_{i=1}^{m}l(g(M(\widetilde{X}_{i})),y_{i})|\\
    \leq&2\widetilde{\mathcal{R}}_{\widetilde{S}}(L_{\mathcal{G}})+C_l\sqrt{\frac{2\log(1/\delta)}{m}}\\
    &|E_{\widetilde{X}\sim\widetilde{\mathcal{D}}}[l(g(M(\widetilde{X})),y)]-\frac{1}{m}\sum\limits_{i=1}^{m}l(g(M(\widetilde{X}_{i})),y_{i})|\\
    \leq&2\widehat{\mathcal{R}}_{\widetilde{S}}(L_{\mathcal{G}})+3C_l\sqrt{\frac{2\log(2/\delta)}{m}}.
\end{array}
\end{equation}

Because $\forall l\in L_{\mathcal{G}}$ is $L_l$-Lipschitz functions, we have
\begin{equation}\label{3.10}
\begin{array}{clc}
    \widehat{\mathcal{R}}_{\widetilde{S}}(L_{\mathcal{G}})\leq\sqrt{2}L_l\widehat{\mathcal{R}}_{\widetilde{S}_{X}}(\mathcal{G}_\mathcal{M})\\
    \widetilde{\mathcal{R}}_{\widetilde{S}}(L_{\mathcal{G}})\leq\sqrt{2}L_l\widetilde{\mathcal{R}}_{\widetilde{S}_{X}}(\mathcal{G}_\mathcal{M}).
\end{array}
\end{equation}
Then,
\begin{equation}\label{3.11}
\begin{array}{clc}
    |R_{\widetilde{\mathcal{D}}}(h)-\widehat{R}_{\widetilde{\mathcal{D}}}(h)|\leq2\sqrt{2}L_l\widetilde{\mathcal{R}}_{\widetilde{S}_{X}}(\mathcal{G}_\mathcal{M})+C_l\sqrt{\frac{2\log(1/\delta)}{m}}\\
    |R_{\widetilde{\mathcal{D}}}(h)-\widehat{R}_{\widetilde{\mathcal{D}}}(h)|\leq2\sqrt{2}L_l\widehat{\mathcal{R}}_{\widetilde{S}_{X}}(\mathcal{G}_\mathcal{M})+3C_l\sqrt{\frac{2\log(2/\delta)}{m}}.\\
\end{array}
\end{equation}
\end{theorem}

The proof of theorem \ref{T-3} is similar to theorem \ref{T-1}.

Next, we consider the estimation error bounds for kernel-based optimization functions such as support vector machine (SVM). Let $K : \mathbb{R}^{p} \times\mathbb{R}^{p} \rightarrow \mathbb{R}$ be a PDS kernel function, $\Phi: \mathbb{R}^{p}\rightarrow \mathbb{H}$ be a feature mapping associated to $K$ and $w_1,\cdots, w_K\in\mathbb{H}$ are weight vectors. For any $p\geq 1$, the family of kernel-based hypotheses is denoted as:
\begin{equation*}
\begin{array}{clc}
    \mathcal{G}_{K,p}=\{g:  M(\widetilde{X})\rightarrow (w_1^{T}\Phi(M(\widetilde{X})),\cdots,w_K^{T}\Phi(M(\widetilde{X}))),\\
    W=(w_1^{T},\cdots,w_K^{T})^{T},||W||_{\mathbb{H},p}\leq \Lambda
    \},
\end{array}
\end{equation*}
where, $||W||_{\mathbb{H},p}=(\sum\limits_{l=1}^{K}{||w_l||_{\mathbb{H}}^{p}})^{1/p}$. Hence, the fuzzy Rademacher complexity of $\mathcal{G}_{K,p}$ can be bounded as follow.

\begin{lemma}\label{L-1}
Let $K : \mathbb{R}^{p} \times\mathbb{R}^{p} \rightarrow \mathbb{R}$ be a PDS kernel function and $\Phi: \mathbb{R}^{p} \rightarrow \mathbb{H}$ be a feature mapping associated to $K$. Assume that there exists $r > 0$ such that $K(M(\widetilde{X}), M(\widetilde{X})) \leq r^2$ for all $\widetilde{X} \in \mathcal{\widetilde{X}}$. Let $\widetilde{S}_{X} = \{\widetilde{X}_{i}\}_{i=1}^m, \widetilde{X}_{i}\sim\widetilde{\mathcal{D}}\in\mathcal{\widetilde{X}}$. Then, for any $m \geq 1$,
\begin{equation}
\begin{array}{clc}
    \widetilde{\mathcal{R}}_{\widetilde{S}_{X}\sim\widetilde{\mathcal{D}}}(\mathcal{G}_{K,p})\leq K\sqrt{\frac{r^{2}\Lambda^{2}}{m}}.
\end{array}
\end{equation}
\begin{proof}
For all $l\in [1,K]$, $||w_l||_{\mathbb{H}}\leq(\sum\limits_{l=1}^{K}{||w_l||_{\mathbb{H}}^{p}})^{1/p}=||W||_{\mathbb{H},p}$ holds. Thus, as $||W||_{\mathbb{H},p}\leq \Lambda$, we have $||w_l||_{\mathbb{H}}\leq \Lambda$ for all $l\in [1,K]$. Then, the fuzzy Rademacher complexity of the hypothesis set $\mathcal{G}_{K,p}$ can be bounded as follows:
\begin{equation*}
\begin{array}{clc}
    &\widetilde{\mathcal{R}}_{\widetilde{S}_{X}\sim\widetilde{\mathcal{D}}}(\mathcal{G}_{K,p})\\
    =&\frac{1}{m}E_{\widetilde{\mathcal{D}},\vec{\sigma}}[\sup\limits_{||W||\leq \Lambda}\sum\limits_{i=1}^{m}\sum\limits_{k=1}^{K}{\sigma_{ik}g_k(M(\widetilde{X}_{i}))}]\\
    =&\frac{1}{m}E_{\widetilde{\mathcal{D}},\vec{\sigma}}[\sup\limits_{||W||\leq \Lambda}\sum\limits_{i=1}^{m}\sum\limits_{k=1}^{K}{\sigma_{ik}w_k^{T}\Phi(M(\widetilde{X}_{i}))}]\\
    \leq&\frac{K}{m}E_{\widetilde{\mathcal{D}},\vec{\sigma}}[\sup\limits_{k\in[K],||W||\leq \Lambda}\langle w_k, \sum\limits_{i=1}^{m}{\sigma_{ik}\Phi(M(\widetilde{X}_{i}))} \rangle]\\
    &(\textnormal{using Cauchy-Schwarz inequality})\\
    \leq&\frac{K}{m}E_{\widetilde{\mathcal{D}},\vec{\sigma}}[\sup\limits_{k\in[K],||W||\leq \Lambda} ||w_k||_{\mathbb{H}} ||\sum\limits_{i=1}^{m}{\sigma_{ik}\Phi(M(\widetilde{X}_{i}))}||_{\mathbb{H}}]\\
    \leq&\frac{K\Lambda}{m}E_{\widetilde{\mathcal{D}},\vec{\sigma}}[\sup\limits_{k\in[K]}  ||\sum\limits_{i=1}^{m}{\sigma_{ik}\Phi(M(\widetilde{X}_{i}))}||_{\mathbb{H}}]\\
    &(\textnormal{using Jensen’s inequality})\\
    \leq&\frac{K\Lambda}{m}[E_{\widetilde{\mathcal{D}},\vec{\sigma}}[\sup\limits_{k\in[K]}  ||\sum\limits_{i=1}^{m}{\sigma_{ik}\Phi(M(\widetilde{X}_{i}))}||^2_{\mathbb{H}}]]^{1/2}\\
    &(i\neq j \Rightarrow E_{\vec{\sigma}}[\sigma_{ik}\sigma_{jk}]=0)\\
    =&\frac{K\Lambda}{m}[E_{\widetilde{\mathcal{D}}}[ \sum\limits_{i=1}^{m}{||\Phi(M(\widetilde{X}_{i}))||^2_{\mathbb{H}}}]]^{1/2}\\
    =&\frac{K\Lambda}{m}[E_{\widetilde{\mathcal{D}}}[ \sum\limits_{i=1}^{m}{K(M(\widetilde{X}_{i}),M(\widetilde{X}_{i}))}]]^{1/2}\\
    \leq& K\sqrt{\frac{r^{2}\Lambda^{2}}{m}},
\end{array}
\end{equation*}
which yields the result.
\end{proof}
\end{lemma}

Next, combining theorem \ref{T-2} and lemma \ref{L-1} directly yields the following generalization bound.
\begin{theorem}\label{T-3}
Let $K : \mathbb{R}^{p} \times\mathbb{R}^{p} \rightarrow \mathbb{R}$ be a PDS kernel function and $\Phi: \mathbb{R}^{p}\rightarrow \mathbb{H}$ be a feature mapping associated to $K$. Assume that there exists $r > 0$ such that $K(M(\widetilde{X}), M(\widetilde{X})) \leq r^2$ for all $\widetilde{X} \in \mathcal{\widetilde{X}}$. Let $\widetilde{S}_{X} = \{\widetilde{X}_{i}\}_{i=1}^m, \widetilde{X}_{i}\sim\widetilde{\mathcal{D}}\in\mathcal{\widetilde{X}}$ and suppose that there are $C, C_l > 0$ such that $\sup_{g\in \mathcal{G}_{K,p}}{\parallel g\parallel_{\infty}}\leq C$ and $\sup_{\parallel g\parallel_{\infty}\leq C}\max_{y}{l(t,y)}\leq C_l$, and $\forall l\in L_{\mathcal{G}_{K,p}}$ is $L_l$-Lipschitz functions. For any $\delta>0$, with fuzzy probability at least $1-\delta$, each of the following holds for all $h\in \mathcal{G}_{K,p}$:
\begin{equation}\label{4.1}
\begin{array}{clc}
    |R_{\widetilde{\mathcal{D}}}(h)-\widehat{R}_{\widetilde{\mathcal{D}}}(h)|\leq2KL_l\sqrt{\frac{2r^{2}\Lambda^{2}}{m}}+C_l\sqrt{\frac{2\log(1/\delta)}{m}}.\\
\end{array}
\end{equation}
\end{theorem}

According to equations (\ref{11}),  (\ref{3.11}), and (\ref{4.1}), we notice that fix some constants, as $m\rightarrow\infty$, $R_{\widetilde{\mathcal{D}}}(h)\rightarrow\widehat{R}_{\widetilde{\mathcal{D}}}(h)$. Therefore, these bounds demonstrate that we can always obtain a fuzzy classifier with high classification accuracy when enough fuzzy-feature instances can be collected. These theoretical analyses reveal that fuzzy classifiers can be constructed to effectively and accurately handle the MCIMO problem.

\section{Construct Fuzzy Classifiers for Solving MCIMO Problem}

\label{sec:algorithm}
\begin{figure}[!t]
\centerline{\includegraphics[scale=0.5]{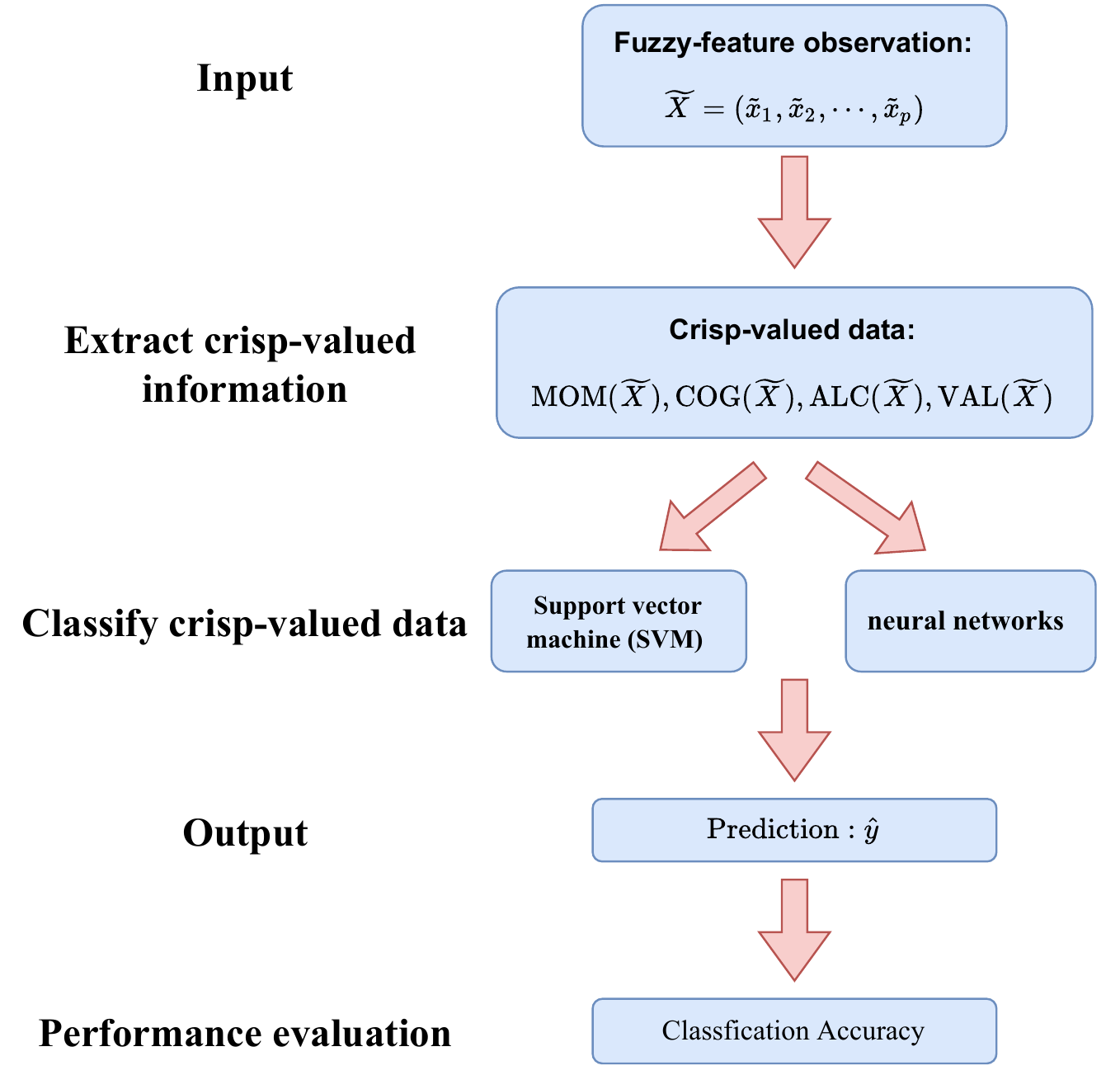}}
\caption{The framework of the proposed algorithms.}
\label{DF}
\end{figure}

In this section, two fuzzy classifiers are constructed to handle the MCIMO problem. The framework of the proposed algorithms is shown in Figure \ref{DF}. In the MCIMO problem, we aim to train a fuzzy classifier for fuzzy-feature input prediction. Let  $\widetilde{X}_{i}=(\widetilde{x}_{i1},\widetilde{x}_{i2},\cdots,\widetilde{x}_{ip}), i=1,\cdots,m$ be a fuzzy-feature input, where $\widetilde{x_{ij}}, i=1,\cdots,m, j=1,\cdots,p$ are fuzzy number. Common used fuzzy numbers include Gaussian fuzzy numbers, trapezoidal fuzzy numbers and triangular fuzzy numbers. Firstly, a Gaussian fuzzy number $\widetilde{x}$ can be characterized by $(c, \delta)$ and the membership function is given in the following equation:
\begin{equation*}
\begin{array}{clc}
     \mu_{\widetilde{x}}(t)=\textnormal{exp}(-(t-c)/2\delta)^2.
\end{array}
\end{equation*}
A trapezoidal fuzzy number $\widetilde{x}$ can be characterized by $(a_1, b_1, b_2, a_2)$ and the membership function of a trapezoidal fuzzy number $\widetilde{x}$ is shown as follow:
\begin{equation*}
    \mu_{\widetilde{x}}(t)=\left\{
             \begin{aligned}
             0,\quad & t<a_1\\
             \frac{t-a_1}{b_1-a_1},\quad & a_1\leq t< b_1\\
             1,\quad & b_1\leq t< b_2\\
             \frac{t-a_2}{b_2-a_2},\quad & b_2\leq t< a_2\\
             0,\quad & t\geq a_2.\\
             \end{aligned}
          \right.
\end{equation*}
Finally, when $b_1 = b_2$, a trapezoidal fuzzy number is become a triangular fuzzy number. Thus, a triangular fuzzy number $\widetilde{x}$ can be characterized by $(a_1, b_1, a_2)$.

To address the MCIMO problem, we need to construct a hypothesis function $h\in \mathcal{H}$ which mapping the input space $\widetilde{\mathcal{X}}\subset\mathcal{F}^{p}_{\mathbb{R}^{p}}$ into $\mathbb{R}^{K}$. A hypothesis function $h$ can be decomposed into a composition of two functions. The first function $M$, called defuzzification function, is defined as follow:
\begin{equation*}
\begin{array}{clc}
     M:& \widetilde{\mathcal{X}}\rightarrow\ \mathbb{R}^{p}\\
    
    & (\widetilde{x}_{i1},\widetilde{x}_{i2},\cdots,\widetilde{x}_{ip})\rightarrow (M(\widetilde{x}_{i1}),\cdots,M(\widetilde{x}_{ip})).
\end{array}
\end{equation*}

Next, four different defuzzification methods are introduced:
\begin{enumerate}
\item The first method is called \emph{Mean/Middle of Maxima} (MOM) \cite{roychowdhury2001survey} which is widely-used due to its calculation simplicity. MOM is defined as:
\begin{equation}\label{6.1}
\begin{array}{clc}
     \textnormal{MOM}(\widetilde{x})=\textnormal{Mean}(t = \arg\max_{t} \mu_{\widetilde{x}}(t)).
\end{array}
\end{equation}
\item \emph{The Centre of Gravity} (COG) \cite{van1999defuzzification} is another widely-used defuzzification method. The definitions of COG for discrete and continuous situations are show
as follow:
\begin{align}
     \textnormal{COG}(\widetilde{x})&=\frac{\sum t \mu_{\widetilde{x}}(t)}{\sum \mu_{\widetilde{x}}(t)}(\textnormal{discrete})\\
     &=\frac{\int t \mu_{\widetilde{x}}(t)dt}{\int \mu_{\widetilde{x}}(t)dt}(\textnormal{continuous}).
\end{align}

\item The third approach, called \emph{averaging
level cuts} (ALC) \cite{oussalah2002compatibility}, is defined as the flat averaging of all midpoints of the $\alpha$-cuts. ALC is defined as :
\begin{equation}\label{6.3}
\begin{array}{clc}
     \textnormal{ALC}(\widetilde{x})=\frac{1}{2}\int_{0}^{1}{(\widetilde{x}_{\alpha}^{L}+\widetilde{x}_{\alpha}^{U})}d\alpha.
\end{array}
\end{equation}
\item The final method is called \emph{value of a fuzzy number} (VAL) \cite{delgado1998canonical} which uses $\alpha$-levels as weighting factors in averaging the $\alpha$-cut midpoints. VAL is defined as :
\begin{equation}\label{6.4}
\begin{array}{clc}
     \textnormal{VAL}(\widetilde{x})=\int_{0}^{1}{\alpha(\widetilde{x}_{\alpha}^{L}+\widetilde{x}_{\alpha}^{U})}d\alpha.
\end{array}
\end{equation}

\end{enumerate}

In Section \ref{sec:experiment}, we compare the performance of different defuzzification methods on synthetic datasets. The experimental results illustrate that VAL outperforms than other three defuzzification methods. Therefore, equation (\ref{6.4}) is used as the defuzzification function in all subsequent experiments.

Through the first progress, the initial issue becomes a traditional multi-class classification problem with crisp data. Therefore, the second function, called the optimization function, is a hypothesis function that maps $\mathbb{R}^{p}$ into $\mathbb{R}^{K}$ to solve the traditional multi-class classification problem. Since support vector machine and neural networks have gained great achievements on multi-classification problems, we decide to apply both algorithms as the optimization method. Next, we will introduce both algorithms for multi-classification problems.

\subsection{Defuzzified support vector machine}
Firstly, support vector machine (one-vs-rest SVM \cite{weston1999support}) with PDS kernel function is used as the optimization function to solve the MCIMO problem.
Suppose $D_{tr} = ((\widetilde{X}_{1},y_{1}),\cdots,(\widetilde{X}_{N},y_{N}))$ is the training data, where $\widetilde{X}_{i}\in\widetilde{\mathcal{X}}\subset\mathcal{F}^{p}_{\mathbb{R}^{p}}, y_{i}\in\{-l,+l\}, l=1,2,\cdots,K, i=1,2,\cdots,N$. The $-l$ indicates that $\widetilde{X}_{i}$ does not belong to category $l$, and the $+l$ represents that $\widetilde{X}_{i}$ belongs to category $l$. In the first step, defuzzification function (\ref{6.4}) is used to transform fuzzy input $\widetilde{D}_{x}=(\widetilde{X}_{1},\cdots,\widetilde{X}_{N})$ to crisp input denoted as $D_{x}=(X_{1},\cdots,X_{N})$. Let $K : \mathcal{X} \times\mathcal{X} \rightarrow \mathbb{R}$ be a PDS kernel function. Hence, we need to solve $K$ optimization problems separately, and the $l$th problem is shown as follows:
\begin{equation}\label{5.1}
\begin{array}{clc}
     &\min\limits_{\alpha} &\frac{1}{2}\sum\limits_{i=1}^{N}\sum\limits_{j=1}^{N}{\alpha_{il}\alpha_{jl}y_{i}y_{j}K(X_{i},X_{j})} - \sum\limits_{i=1}^{N}\alpha_{il}\\
     &s.t &\sum\limits_{i=1}^{N}\alpha_{il}y_{i}=0\\
     &  &0\leq \alpha_{il} \leq C, i=1,2,\cdots,N.
\end{array}
\end{equation}

The optimal solution is $\overrightarrow{\alpha}^{*}_{l}=(\alpha^{*}_{1l},\cdots,\alpha^{*}_{Nl})^{T}, l=1,2,\cdots,K$. Then, choose a positive component $0\leq\alpha^{*}_{jl}\leq C$ of $\overrightarrow{\alpha}^{*}_{l}$, and calculate
\begin{equation}\label{5.2}
\begin{array}{clc}
     b^{*}_{l}=y_{j}-\sum\limits_{i=1}^{N}\alpha^{*}_{il}y_{i}K(X_{i},X_{j}).
\end{array}
\end{equation}

Finally, the decision function is given as follow:
\begin{equation}\label{5.3}
\begin{array}{clc}
     h(X)=\arg\max_{l\in[K]}(\sum\limits_{i=1}^{N}\alpha^{*}_{il}y_{i}K(X,X_{i})+b^{*}_{l}).
\end{array}
\end{equation}

The following algorithm called \emph{defuzzified support vector machine} (DF-SVM) is shown in Algorithm \ref{A1}.

\begin{algorithm}[t]
\small
{\bfseries 1: Input} training data ${D}_{tr}$, selected appropriate regularization parameter $C$ and kernel function ;

{\bfseries 2: Initial}  Preprocessing the training data ${D}_{tr}$;

{\bfseries 3: Defuzzification} Using equation (\ref{6.4}) to transform $\widetilde{D}_{x}=(\widetilde{X}_{1},\cdots,\widetilde{X}_{N})$ into $D_{x}=(X_{1},\cdots,X_{N})$;

{\bfseries 4: Optimization} \\
Solving $K$ optimization problems in (\ref{5.1});

{\bfseries 5: Output} $\overrightarrow{\alpha}^{*}_{l}=(\alpha^{*}_{1l},\cdots,\alpha^{*}_{Nl})^{T}, l=1,2,\cdots,K$ and the decision function in (\ref{5.3}).
\caption{DF-SVM}\label{A1}
\end{algorithm}

\subsection{Defuzzified multilayer perception}

Secondly, a multilayer perception model, which contains two hidden layers and an output layer (softmax), is used as the optimization function to complete the second progress. We denote the parameters of the two hidden layers are $W_1,b_1$ and $W_2,b_2$ respectively, and the parameters of the output layer are $W_0,b_0$ respectively, and the activation function is $\phi$. Then, the outcome of the constructed multilayer perception model can be expressed as when we get a fuzzy-feature input $\widetilde{X}$:
\begin{equation}\label{3.2}
\begin{array}{clc}
     O(\widetilde{X})=\phi(\phi(M(\widetilde{X})W_1+b_1)W_2+b_2)W_0+b_0,\\
     \widehat{y}=\arg\max_{k\in \{1,2,\cdots,K\}}(h_k(\widetilde{X})),
\end{array}
\end{equation}
where
\begin{equation*}
\begin{array}{clc}
    h(\widetilde{X})=(h_1(\widetilde{X}),\cdots,h_K(\widetilde{X}))=\textnormal{softmax}(O(\widetilde{X})).
\end{array}
\end{equation*}

The following algorithm called \emph{defuzzified multilayer perception} (DF-MLP) is shown in Algorithm \ref{A2}.

\begin{algorithm}[t]
\small
{\bfseries 1: Input} training data ${D}_{tr}$, learning rate $\eta$, fixed epoch $T_{max}$, loss function (cross-entropy loss function is selected) and optimization algorithm (Adam algorithm \cite{kingma2015adam} is selected);

{\bfseries 2: Initial}   $W_0^0$, $W_1^0$, $W_2^0$, $b_0^0$, $b_1^0$, $b_2^0$;

\For{$T = 1,2,\dots,T_{max}$}{

{\bfseries 3: Fetch} mini-batch $\check{D}_{tr}$ from ${D}_{tr}$;

{\bfseries 4: Calculate} \\
$L = loss(h(\widetilde{X};W_0^{T-1}, W_1^{T-1}, W_2^{T-1}, b_0^{T-1}, b_1^{T-1}, b_2^{T-1}),\widehat{y})$\\ according to Eqs.~(\ref{6.4}) and (\ref{3.2});

{\bfseries 5: Update} $W_0^{T}$, $W_1^{T}$, $W_2^{T}$, $b_0^{T}$, $b_1^{T}$, $b_2^{T} = \textnormal{Adam}(L)$;

}

{\bfseries 6: Output} $W_0^{T_{max}}$, $W_1^{T_{max}}$, $W_2^{T_{max}}$, $b_0^{T_{max}}$, $b_1^{T_{max}}$, $b_2^{T_{max}}$.
\caption{DF-MLP \cite{ma2021learning}}\label{A2}
\end{algorithm}

\section{Experiments on Synthetic Datasets}
\label{sec:experiment}
In this section, we first compare the performance of different defuzzification methods on synthetic datasets to select the optimal defuzzification function for the proposed algorithms. Then, we verify the efficacy of the proposed algorithms for solving the MCIMO problem by comparing seven baselines in terms of classification accuracy on synthetic datasets.

\subsection{Dataset generation}
In this section, we introduce how to construct the synthetic dataset (Balanced data) which contains $N$ fuzzy instances distributed in five categories. Each instance has $20$ fuzzy features. Firstly, we generate the real-valued vectors $X_i=(x_{i1},\cdots,x_{i20}),i=1,\cdots,N$ in five categories by a random number generator as the true value of the instance. Then, we use the generated real-valued vectors to construct the observation datasets $\{\widetilde{X}_{i}=(\widetilde{x}_{i1},\cdots,\widetilde{x}_{i20})\}_{i=1}^{N}$. Each $\widetilde{x}_{ij}$ is a triangular fuzzy number characterized by $(x_{ij}-a_{ij}, x_{ij}+b_{ij}, x_{ij}+c_{ij})$ where $a_{ij}\sim U[1.5, 3],b_{ij}\sim U[-0.5, 0.5],c_{ij}\sim U[2, 4]$ and $U[a,b]$ denotes the uniform distribution over $[a,b]$. 

\subsection{Experimental setup}
In this section, baselines and experimental details of all baselines, DF-SVM and DF-MLP are introduced.
\subsubsection{Baselines}
Firstly, we introduce the first five baselines which called Meanlogistic, MeanSVM, MeanDecisiontree, MeanRandomForest and MeanMLP. For fuzzy-feature dataset, a fuzzy feature is denoted as $\widetilde{x}=(\inf P_0, \sup P_0, \inf P_1, \sup P_1)$. We use $M_{1}(\widetilde{x})=(\inf P_0 +\sup P_0+ \inf P_1+ \sup P_1)/4$ to transfer fuzzy features to crisp features. For interval-valued datasets, $x=[A,B]$ is denoted as a interval-valued feature. Similarly, $M_{2}(x)=(A+B)/2$ is used to transfer interval-valued features to crisp features. Then, those baselines apply five well-known machine learning methods (logistic regression, SVM, decision trees, random forests and neural networks) to classify crisp-valued data obtained with the above-mentioned methods. Secondly, the last two baselines called DCCF and BCCF are presented in \cite{Colubi2011Nonparametric}. 

\begin{table*}[t!]
\centering
\caption{Hyperparameters for the proposed algorithms and seven baselines}
\label{table_00}
\renewcommand{\arraystretch}{1.5}
\begin{tabular}{ccc}
\hline
Algorithm  & Hyperparameters & Ranges\\ \hline
Meanlogistic  & regularization parameter $C$ & $\{0.1, 0.2, \cdots, 0.9, 1, 2, \cdots,100\}$\\ \hline
MeanSVM &  regularization parameter $C$, kernel type & $\{0.1, 0.2, \cdots, 0.9, 1, 2, \cdots,100\}$, \{‘linear’, ‘poly’, ‘rbf’\}\\ \hline
MeanDecisiontree & min samples leaf & $\{1, 2, \cdots, 10\}$\\ \hline
MeanRandomForest & min samples leaf, the number of trees & $\{1, 2, \cdots, 10\}$, $\{5, 10, \cdots, 100\}$\\ \hline
MeanMLP & learning rate, hidden layer units, epochs & $\{0.0001, 0.001, 0.01, 0.1\}, \{20, 30, \cdots, 200\}, \{100, 200, 500, 1000, 1500\}$\\ \hline
DCCF\cite{Colubi2011Nonparametric} & bandwidth $h_g$ & $\{1, 2, \cdots,10, 20, \cdots,50\}$\\ \hline
BCCF\cite{Colubi2011Nonparametric} & distance parameter $\delta$ & $\{0.1, 0.5,  1, 2, \cdots,10\}$\\ \hline
DF-SVM & regularization parameter $C$, kernel type & $\{0.1, 0.2, \cdots, 0.9, 1, 2, \cdots,100\}$, \{‘linear’, ‘poly’, ‘rbf’\}\\ \hline
DF-MLP & learning rate, hidden layer units, epochs & $\{0.0001, 0.001, 0.01, 0.1\}, \{20, 30, \cdots, 200\}, \{100, 200, 500, 1000, 1500\}$\\ \hline
\end{tabular}
\end{table*}

\begin{figure*}[tp]
    \begin{center}
        \subfigure[DF-SVM with $4$ defuzzification functions.]
        {\includegraphics[width=0.32\textwidth]{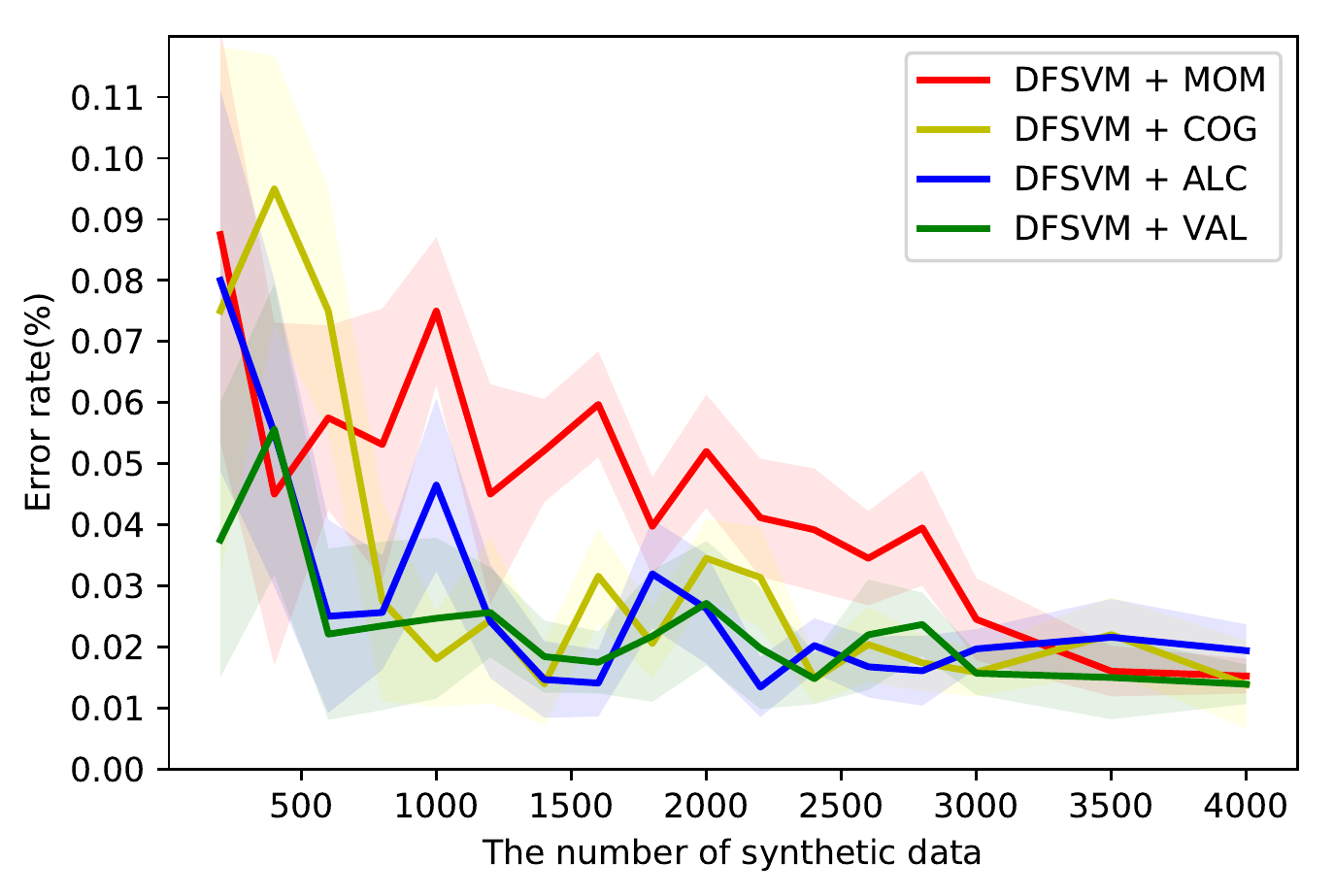}\label{fig_svm_case}}
        \subfigure[DF-MLP with $4$ defuzzification functions.]
        {\includegraphics[width=0.32\textwidth]{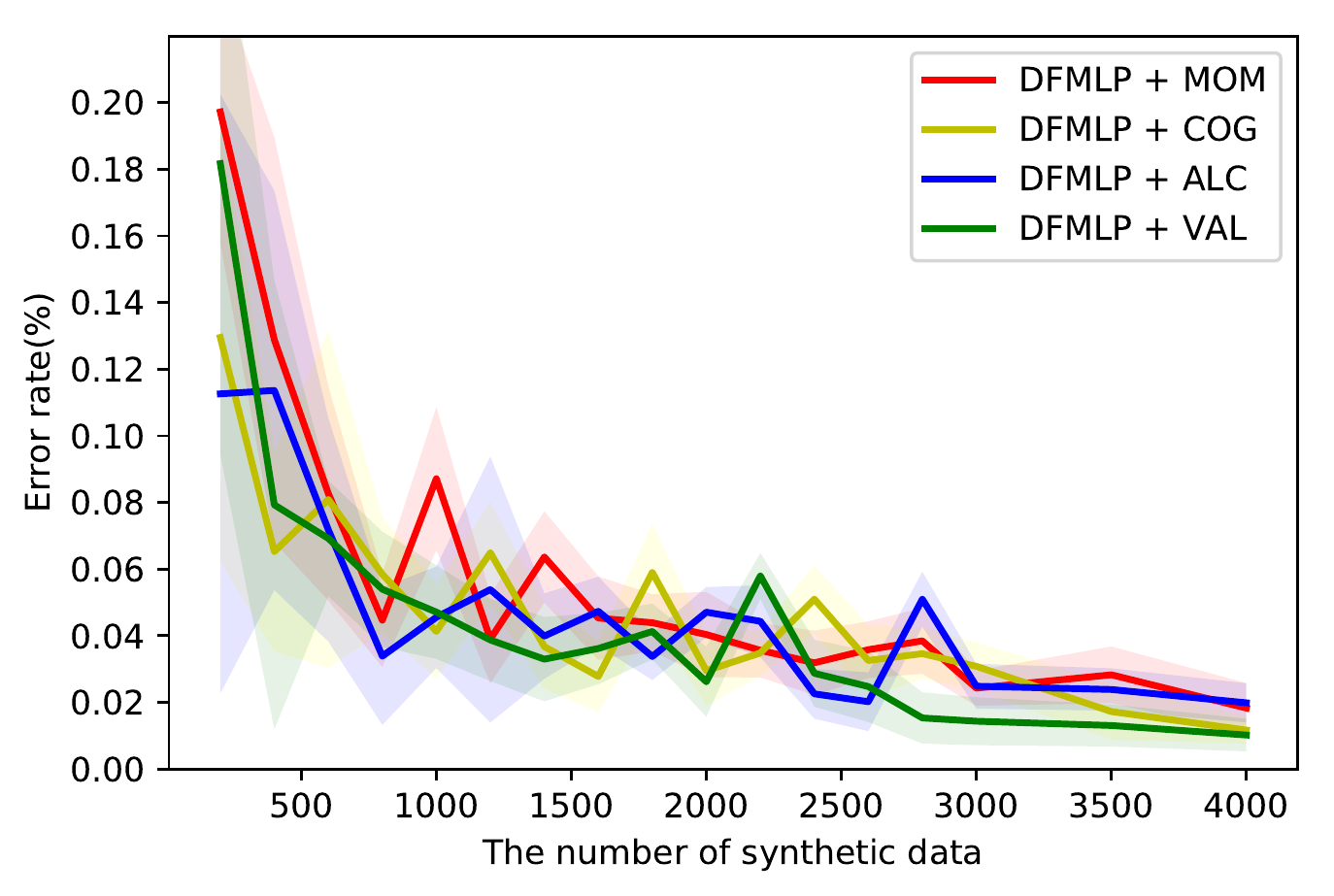}\label{fig_mlp_case}}
        \subfigure[DF-SVM and DF-MLP with VAL.]
        {\includegraphics[width=0.32\textwidth]{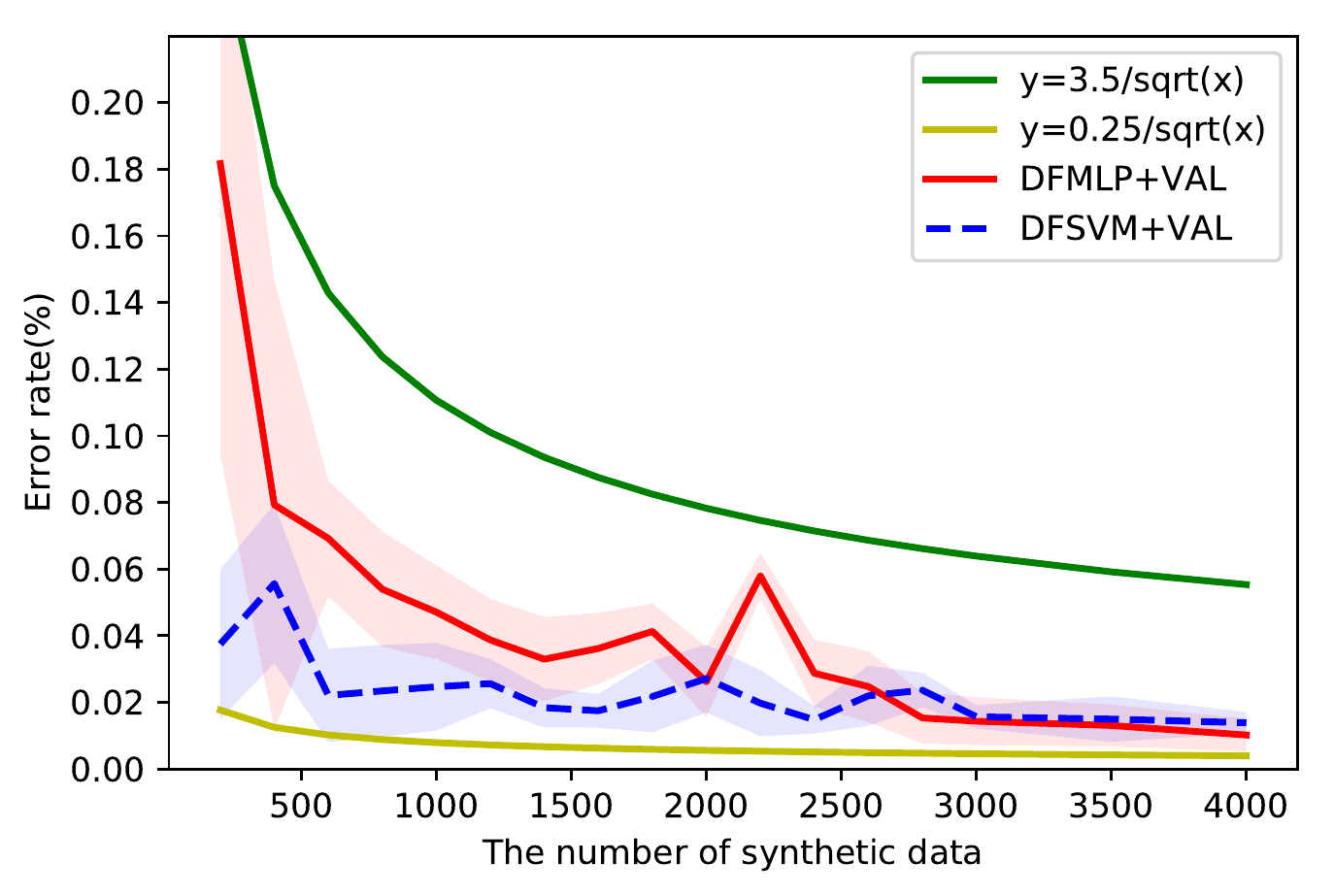}\label{fig_val_case}}
        \caption{Classification error rate on the test set varies with the number of synthetic data. 
        }\label{fig001}
    \end{center}
\end{figure*}

\begin{figure}[!t]
\centerline{\includegraphics[scale=0.55]{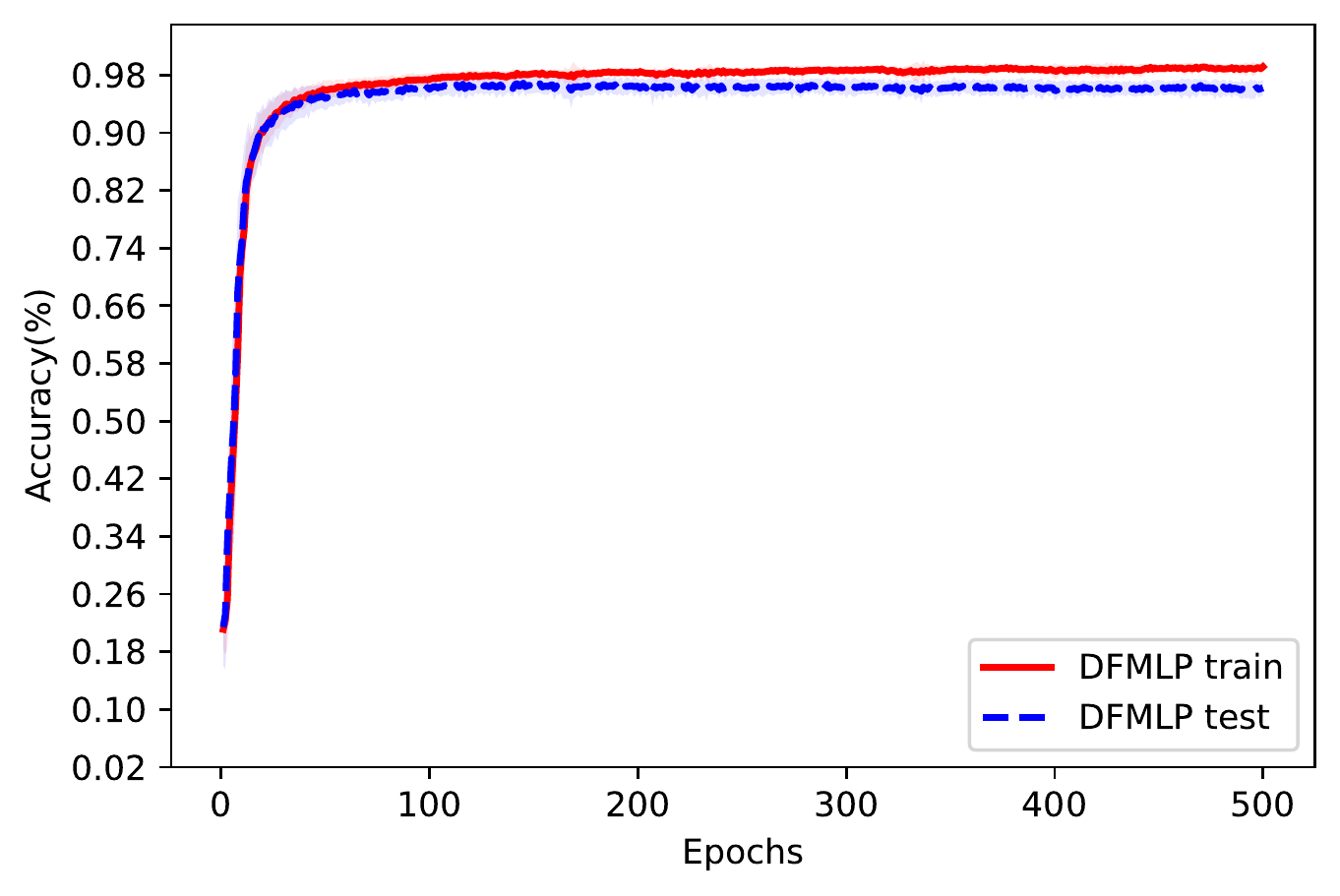}}
\caption{Accuracy curve on the synthetic datasets vs. the number of epochs.}
\label{fig_sys}
\end{figure}

\subsubsection{Experimental details}
For DF-MLP, we let $\textnormal{momentum}=0.9$ and $\textnormal{weight decay}=0.0001$. Finally, for the DCCF and BCCF algorithms, $\varphi$ is selected to be the Lebesgue measure on $[0,1]$ and $\theta=1/3$, $K(u)=\frac{15}{8}(1-u^2)^{2}I_{(u\in[0,1])}$ is used as the kernel function. All these settings of DCCF and BCCF algorithms can obtain the best performance from \cite{Colubi2011Nonparametric}. However, DCCF and BCCF algorithms can only process the fuzzy data with one fuzzy feature, whereas the generated synthetic datasets contain multiple fuzzy features. Therefore, we consider using the average distance between each fuzzy feature to represent the distance between the fuzzy feature vectors in the DCCF and BCCF algorithms.

For each algorithm on each dataset, we randomly divide each dataset into the training set, the validation set and the test set, which contain $60\%$, $20\%$ and $20\%$ of the data, respectively. First, we select the hyperparameters that can obtain the highest average classification accuracy on the validation set. The average classification accuracy on the validation set is the average of the results of $10$ repeated experiments on the validation set. The hyperparameters that need to be selected are shown in Table \ref{table_00}. Then, the selected optimal hyperparameters are used to test the performance of each algorithm on the test set. We repeat the entire experiment process $20$ times. Thus, the final results are shown in the form of "mean$\pm$ standard deviation." To avoid random errors, we randomly scramble the data before each experiment. Classification accuracy is used to evaluate the performance of the proposed model. The definition of classification accuracy is shown as follows:
\begin{equation*}
    {\rm Accuracy} = \frac{|\widetilde{X} \in \mathcal{\widetilde{X}}:f(\widetilde{X})=h(\widetilde{X})|}{|\widetilde{X} \in \mathcal{\widetilde{X}}|},
\end{equation*}
where $f(\widetilde{X})$ is the ground truth label of $\widetilde{X}$, while $h(\widetilde{X})$ is the label predicted by the presented algorithms and the baselines.

In the first experiment, we compare the performance of the proposed two algorithms with different defuzzification functions on the test set when the number of synthetic data increases. The number of synthetic data $N$ is selected from $\{200, 400, \cdots, 3000, 3500, 4000\}$.
In the second experiment, we generated $2000$ synthetic data and analyzed them using the proposed methods and baselines, respectively. In addition, the Wilcoxon rank-sum test results of the method, which obtains the best performance, with other methods are given.

\subsection{Experimental results analysis}
The results of the first experiment are shown in Figure \ref{fig001}. From Figures \ref{fig_svm_case} and \ref{fig_mlp_case}, we find that COG and VAL have better performance than another two methods in terms of convergence speed and classification error and VAL is more stable than the other three methods. The reason why VAL can achieve better performance than other methods is that VAL uses all information from fuzzy sets so that some key information is not discarded. In addition, VAL gives less importance to the lower levels of fuzzy sets, which is reasonable from the perspective of the concept of membership function. Therefore, we use VAL as the defuzzification method in the following experiments. Moreover, from Figure \ref{fig_val_case}, it illustrates that the convergence rate of the two proposed algorithms with VAL defuzzification method is $O(1/\sqrt{m})$. Therefore, we confirmed the theoretical analysis results in Section \ref{sec:Theoretical Analysis} that we can always obtain a fuzzy classifier with high classification accuracy when sufficient fuzzy-feature observations are available.

The results of the second experiment are illustrated in Table \ref{table_0}, and Figure \ref{fig_sys} shows the classification accuracy curve of Algorithm \ref{A2} on the synthetic datasets vs. the number of epochs. From the results, DF-SVM and DF-MLP obtain better performance than the most other baselines on the synthetic dataset. Further, the results of the statistic test show that DF-SVM outperforms other methods significantly at the $0.05$ significance level, which demonstrates the superiority of the proposed algorithms. In addition, we present the experimental running times for the proposed algorithms and all baselines.

\begin{table}[!t]
\centering
\caption{Experiment Result of Synthetic Dataset.}
\label{table_0}
\renewcommand{\arraystretch}{1.5}
\begin{tabular}{p{2cm}<{\centering}p{2.1cm}<{\centering}p{1.6cm}<{\centering}p{1.5cm}<{\centering}}
\hline
Algorithms  & Test accuracy & p & Time (sec)\\ \hline
Meanlogistic  & 96.86$\%$ $\pm$0.87$\%$ & ${2.2\times 10^{-6}} ^{\ast}$ & 119.97\\ \hline
MeanSVM &  97.72$\%$ $\pm$0.71$\%$ & $0.0337^{\ast}$ & 127.35 \\ \hline
MeanDecisiontree & 78.20$\%$ $\pm$2.70$\%$ & ${6.3\times 10^{-8}}^{\ast}$ & 2.23\\ \hline
MeanRandomForest & 95.82$\%$ $\pm$0.85$\%$ & ${9.8\times 10^{-8}}^{\ast}$ & 1088.57\\ \hline
MeanMLP & 96.16$\%$ $\pm$0.80$\%$ & ${3.7\times 10^{-7}}^{\ast}$ & 6607.89\\ \hline
DCCF\cite{Colubi2011Nonparametric} & 92.58$\%$ $\pm$1.02$\%$ &  ${6.3\times 10^{-8}}^{\ast}$ & 1122687\\ \hline
BCCF\cite{Colubi2011Nonparametric} & 92.51$\%$ $\pm$ 1.03$\%$ & ${6.3\times 10^{-8}}^{\ast}$ & 1123543\\ \hline
DF-SVM & $\mathbf{98.24\%\pm0.52\%}$ & ---- & 119.98\\ \hline
DF-MLP & 96.90$\%$ $\pm$ 0.95$\%$& ${2.2\times 10^{-5}}^{\ast}$ & 6593.64\\ \hline
\end{tabular}
\begin{tablenotes}
\item[1] The bold value represents the highest accuracy in each column.
\item[2] p: The p-value of the Wilcoxon rank-sum test between the performance of DF-SVM and other algorithms.
\item[3] $^{\ast} p<0.05$ 
\end{tablenotes}
\small
\end{table}

\section{Experiments on Real-World Datasets}
\label{sec:experiments}
In this section, five real-world datasets are used to verify the efficacy of proposed algorithms for solving the MCIMO problem by comparing with seven baselines in terms of classification accuracy. Besides, we show how to apply the proposed algorithms to analyze interval-valued datasets.

\subsection{Real-world datasets}
In this section, we briefly introduce the five real-world datasets used in the experiments.

\subsubsection{Perceptions experiment dataset}
The $1$st dataset, called the perceptions experiment dataset, contains 551 observations with one fuzzy feature. The fuzzy feature is a trapezoidal fuzzy number characterized by $(\inf P_0, \sup P_0, \inf P_1, \sup P_1)$. Each observation is the perceptions experiment result for one person. The description of perceptions experiment can be found in the following URL: http://bellman.ciencias.uniovi.es/SMIRE/Perceptions.html.

\begin{figure}[!t]
\centerline{\includegraphics[scale=0.45]{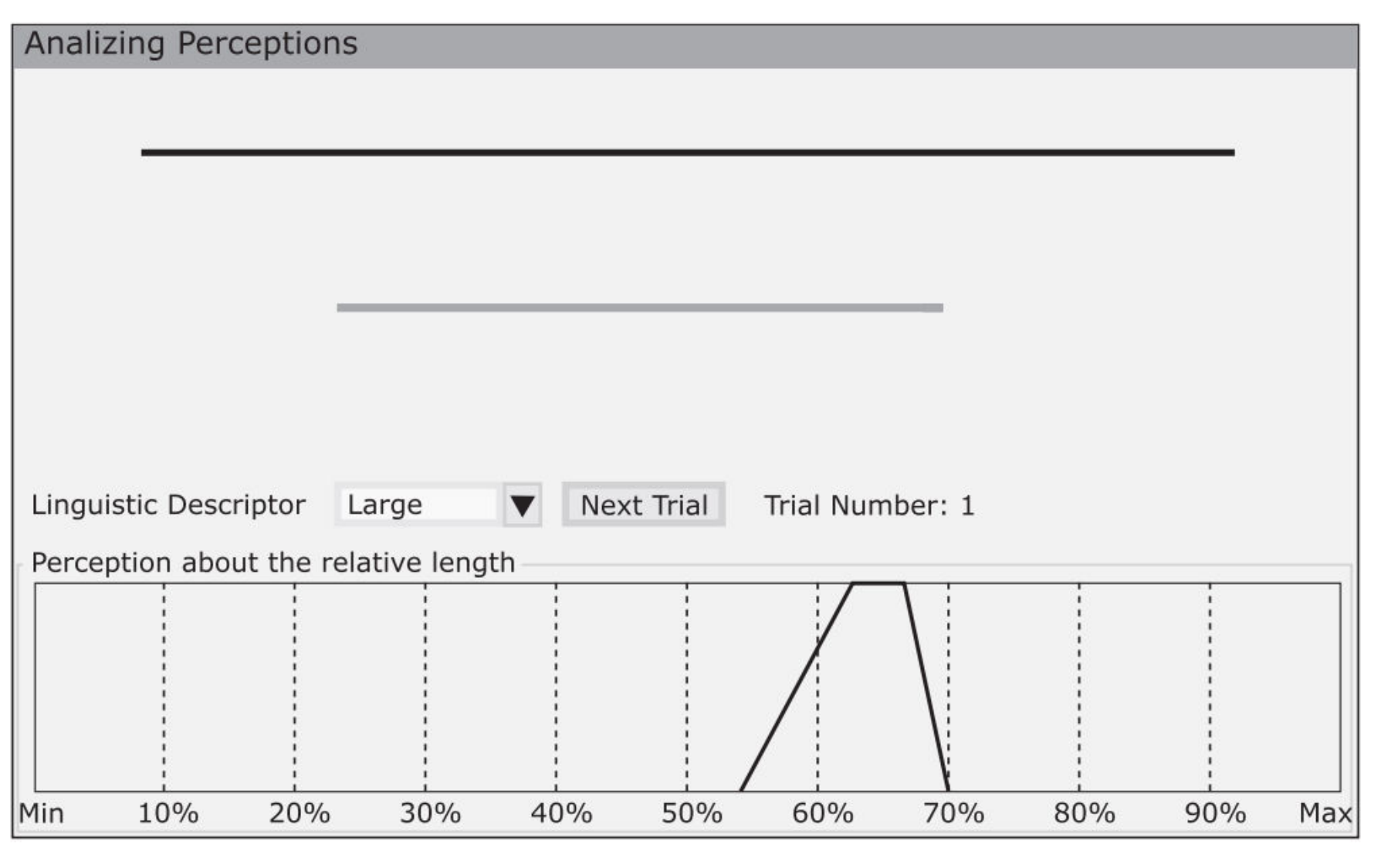}}
\caption{Software to evaluate the visual perception of a line segment.}
\label{fig1}
\end{figure}

In the perceptions experiment, the one black line that people will see is shown in Figure \ref{fig1}. Once participants see a black line, they will be asked to give a trapezoidal fuzzy number characterized by $(\inf P_0, \sup P_0, \inf P_1, \sup P_1)$ to describe it.

For the first dataset, we consider using the fuzzy feature (i.e., the trapezoidal fuzzy number) to predict the category (very small; small; medium; large or very large), which will be selected by the participants according to their perception of the black line.

\subsubsection{Mushroom dataset}
The $2$nd dataset is the California mushroom dataset\footnote{See https://www.mykoweb.com/CAF/ for more details.} that contains $245$ instances in $17$ fungi species categories. There are five interval-valued variables: the pileus cap width ($X_1$), the stipe length ($X_2$), the stipe thickness ($X_3$), the spores major axis length ($X_4$), and the spores minor axis length ($X_5$). Some instances of the mushroom dataset are shown in Table \ref{table_1}. The goal of our experiment on this dataset is to predict the species category of the California mushroom using five interval-valued features. 

\subsubsection{Letter Recognition dataset}
The $3$rd dataset is the letter recognition dataset, selected from UCI Machine Learning Repository (https://archive-beta.ics.uci.edu/), which contains $20000$ instances in $26$ categories. This dataset contains $16$ integer features extracted from raster scan images of the letters. We use the same methods described in Section \ref{sec:experiment} to transfer integer features into fuzzy features. Then, we obtain one real-world dataset with fuzzy-valued features. The goal of our experiment on this dataset is to identify each of a large number of black-and-white rectangular pixel displays as one of the 26 capital letters in the English alphabet. 

\subsubsection{London weather dataset} 
The $4$th dataset is the meteorological data of London (from March 1, 2016 to December 31, 2021), provided by the ‘Reliable Prognosis’ site (https://rp5.ru/), which contains $2131$ instances. Each instance is meteorological data of one day in London, which described by five interval-valued variables (air temperature $T$, atmospheric pressure at weather station level $P0$, atmospheric pressure reduced to main sea level $P$, humidity $U$ and dew-point temperature $Td$) and one category variable (Precipitation or not: $0\equiv $ No Precipitation, $1\equiv $ Precipitation). Some instances of this dataset are shown in Table \ref{table_wea}. We aim to use the five interval-valued features for precipitation prediction.

\subsubsection{Washington weather dataset} 
The $5$th dataset is the meteorological data of Washington (from January 1, 2016 to December 31, 2021) in the ‘Reliable Prognosis’ site as well, which contains $2191$ instances. Each instance is meteorological data of one day in Washington, which described by five interval-valued variables (same as the $4$th dataset) and one category variable (same as the $4$th dataset). We aim to use the five interval-valued features for precipitation prediction.

\begin{table}[!t]
\centering
\caption{Some Instances of the Mushroom Dataset}
\label{table_1}
\renewcommand{\arraystretch}{1.5}
\begin{tabular}{cccccc}
\hline
Species & $X_1$(cm) & $X_2$(cm) & $X_3$(cm) & $X_4$(cm) & $X_5$($\mu$m) \\ \hline
Agaricus & [6,12] & [2,7] & [1.5,3] & [6,7.5] & [4,5] \\ \hline
Boletus & [7,14] & [5,9] & [3,6] & [11.5,13.5] & [3.5,4.5] \\ \hline
Amanita & [6,12] & [9,17] & [1,2] & [9.5,11.5] & [8.5,10] \\ \hline
Clitocybe & [2,9] & [2,6] & [0.5,1.2] & [5,6] & [2.5,3.5] \\ \hline
\end{tabular}
\end{table}

\begin{table}[!t]\scriptsize
\centering
\caption{Some Instances of the London Weather Data}
\label{table_wea}
\renewcommand{\arraystretch}{1.5}
\begin{tabular}{p{0.9cm}<{\centering}p{1cm}<{\centering}p{1.1cm}<{\centering}p{1cm}<{\centering}p{0.6cm}<{\centering}p{1.1cm}<{\centering}p{0.2cm}<{\centering}}
\hline
Times & T & P0 & P & U & Td & Y\\ \hline
31/12/2021 & [0.8,6.1] & [730.2,733.4] & [755.5,759] & [76,99] & [0,3.3] & 1\\ \hline
30/12/2021 & [-1.4,1.5] & [734.2,735.8] & [759.8,762] & [77,93] & [-2.4,-0.6] & 0\\ \hline
29/12/2021 & [-1.2,2.1] & [730.5,735.4] & [756,761] & [93,97] & [-2.4,1.7] & 1\\ \hline
28/12/2021 & [-1.2,1.4] & [730.5,734.2] & [756.1,760] & [72,96] & [-4.2,0.1] & 1\\ \hline
\end{tabular}
\end{table}

\subsection{Preprocessing of interval-valued data}
We notice that the features of the $2$nd, $4$th and $5$th datasets are interval-valued. Therefore, in this section, we present an approach to transform interval-valued features into fuzzy-valued features. Suppose $[A,B]$ is denoted as a feature of one interval-valued instance. Thus, we use one approach that maps $[A,B]$ to a triangular fuzzy number $\widetilde{x}$ characterized by $(A, \beta A + (1-\beta)B, B)$, where $\beta\in [0,1]$ is a hyperparameter to control the shape of the membership function of $\widetilde{x}$. 

Through the above preprocessing, the DF-SVM and DF-MLP algorithms can be used to classify dataset with interval-valued instances. In addition, we realize that the second dataset is an imbalanced dataset which means that each category contains a different number of instances. Therefore, a random oversampling technique (KMeansSMOTE \cite{last2017oversampling}) is used to improve the performance of the proposed algorithms. After the process of the random oversampling technique, the data of each category in the second dataset is expanded to $30$.

\subsection{Experimental setup}
We use the same baselines in Section \ref{sec:experiment}, and the experimental details of all methods are basically the same as in Section \ref{sec:experiment}. The only difference is that one more hyperparameter $\beta$ needs to be selected when analyzing the second dataset. We select the shape parameter $\beta$ from $\{0, 0.05, 0.1, \cdots ,1\}$. Further, we complete the Wilcoxon rank-sum tests of the method, which obtains the best performance, with other methods on real-world datasets. Since DCCF and BCCF can not well handle the dataset with a large number of instances, we only compare the proposed algorithms with the first five baselines on the last three datasets in our experiments.

In addition, since the second dataset is an imbalanced dataset, we use balanced accuracy \cite{brodersen2010balanced} and $AUC$ instead of classification accuracy to compare model performance on the second dataset. The definition of balanced accuracy is
\begin{equation*}
\begin{array}{clc}
    \textnormal{Balanced Accuracy} = \frac{1}{K}\sum\limits_{k=1}^K{(\textnormal{Recall of $k$-th class)}},\\
    \textnormal{Recall} = \rm TP/(TP + FN),
\end{array}
\end{equation*}
where TP is true positive, TN is true negative, FP is false positive and FN is false negative. $AUC$ is equal to the compute area under the receiver operating characteristic curve.

\begin{table}[!t]
\centering
\caption{Experiment Result of Perceptions Experiment Dataset. }
\label{table_2}
\renewcommand{\arraystretch}{1.5}
\begin{tabular}{p{2.5cm}<{\centering}p{2.5cm}<{\centering}p{2.5cm}<{\centering}}
\hline
Algorithms  & Test accuracy & p\\ \hline
Meanlogistic  & 90.04$\%$ $\pm$2.20$\%$ & $0.0080^{\ast}$\\ \hline
MeanSVM &  90.36$\%$ $\pm$2.98$\%$ & 0.5075\\ \hline
MeanDecisiontree & 89.32$\%$ $\pm$3.30$\%$ & $0.0231^{\ast}$\\ \hline
MeanRandomForest & 90.27$\%$ $\pm$3.10$\%$ & 0.3169\\ \hline
MeanMLP & 90.45$\%$ $\pm$2.91$\%$ & 0.3793\\ \hline
DCCF\cite{Colubi2011Nonparametric} & 87.82$\%$ $\pm$2.15$\%$ & $0.0001^{\ast}$\\ \hline
BCCF\cite{Colubi2011Nonparametric} & 88.23$\%$ $\pm$ 2.01$\%$ & $0.0001^{\ast}$\\ \hline
DF-SVM & 91.00$\%$ $\pm$2.52$\%$ & 0.7251\\ \hline
DF-MLP & $\mathbf{91.50\%\pm2.51\%}$ & ----\\ \hline
\end{tabular}
\begin{tablenotes}
\item[1] The bold value represents the highest accuracy in each column.
\item[2] p: The p-value of the Wilcoxon rank-sum test between the performance of DF-MLP and other algorithms.
\item[3] $^{\ast} p<0.05$ 
\end{tablenotes}
\small
\end{table}

\begin{table}[!t]
\centering
\caption{Experiment Result of Mushroom Dataset.}
\label{table_4}
\renewcommand{\arraystretch}{1.5}
\begin{tabular}{p{2.5cm}<{\centering}p{2.5cm}<{\centering}p{2.5cm}<{\centering}}
\hline
Algorithms & Balanced accuracy & AUC\\ \hline
Meanlogistic  & 71.36$\%$ $\pm$3.86$\%$  &  0.9645 $\pm$ 0.0079\\ \hline
MeanSVM &  79.08$\%$ $\pm$3.08$\%$   &  0.9728 $\pm$ 0.0071\\ \hline
MeanDecisiontree & 70.68$\%$ $\pm$4.16$\%$  & 0.9069 $\pm$ 0.0203\\ \hline
MeanRandomForest & 79.04$\%$ $\pm$3.83$\%$  & 0.9750 $\pm$ 0.0077\\ \hline
MeanMLP & 80.49$\%$ $\pm$3.40$\%$  & 0.9721 $\pm$ 0.0071\\ \hline
DCCF\cite{Colubi2011Nonparametric} & 65.14$\%$ $\pm$5.31$\%$ & 0.9584 $\pm$ 0.0078\\ \hline
BCCF\cite{Colubi2011Nonparametric} & 64.16$\%$ $\pm$4.53$\%$ & 0.9554 $\pm$ 0.0083 \\ \hline
DF-SVM & 81.71$\%$ $\pm$4.44$\%$ & 0.9758 $\pm$ 0.0103\\ \hline
DF-MLP & $\mathbf{83.57\%\pm2.04\%}$ & $\mathbf{0.9784 \pm0.0025 }$\\ \hline
\end{tabular}
\begin{tablenotes}
\item[1] The bold value represents the highest accuracy in each column.
\end{tablenotes}
\small
\end{table}

\begin{table}[!t]
\centering
\caption{The p-value of the Statistic Test on Mushroom Dataset.}
\label{table_test}
\renewcommand{\arraystretch}{1.5}
\begin{tabular}{p{3.8cm}<{\centering}p{2.2cm}<{\centering}p{1.8cm}<{\centering}}
\hline
Algorithms & Balanced accuracy & AUC\\ \hline
DF-MLP vs Meanlogistic  & ${6.3\times10^{-8}}^{\ast}$  &  $0.0012^{\ast}$\\ \hline
DF-MLP vs MeanSVM &  ${3.5\times10^{-5}}^{\ast}$   &  0.4171\\ \hline
DF-MLP vs MeanDecisiontree & ${6.3\times10^{-8}}^{\ast}$  & ${6.3\times10^{-8}}^{\ast}$\\ \hline
DF-MLP vs MeanRandomForest & $0.0002^{\ast}$  & 0.0935\\ \hline
DF-MLP vs MeanMLP & $0.0041^{\ast}$  & 0.6849\\ \hline
DF-MLP vs DCCF\cite{Colubi2011Nonparametric} & ${6.3\times10^{-8}}^{\ast}$ & ${6.2\times10^{-5}}^{\ast}$\\ \hline
DF-MLP vs BCCF\cite{Colubi2011Nonparametric} & ${6.3\times10^{-8}}^{\ast}$ & ${2.5\times10^{-5}}^{\ast}$ \\ \hline
DF-MLP vs DF-SVM & 0.1762 & 0.3438 \\ \hline
\end{tabular}
\begin{tablenotes}
\item[1] $^{\ast} p<0.05$ 
\end{tablenotes}
\small
\end{table}

\begin{table}[!t]
\centering
\caption{Experiment Result of Letter Recognition Dataset.}
\label{table_letter}
\renewcommand{\arraystretch}{1.5}
\begin{tabular}{p{2.5cm}<{\centering}p{2.5cm}<{\centering}p{2.5cm}<{\centering}}
\hline
Algorithms & Test accuracy & p\\ \hline
Meanlogistic  & 73.50$\%$ $\pm$0.70$\%$  &  ${6.3\times10^{-8}}^{\ast}$\\ \hline
MeanSVM &  94.60$\%$ $\pm$0.36$\%$   & $0.0011^{\ast}$ \\ \hline
MeanDecisiontree & 78.09$\%$ $\pm$0.69$\%$ & ${6.3\times10^{-8}}^{\ast}$ \\ \hline
MeanRandomForest & 93.50$\%$ $\pm$0.41$\%$ & ${6.3\times10^{-8}}^{\ast}$ \\ \hline
MeanMLP & 91.79$\%$ $\pm$0.47$\%$ & ${6.3\times10^{-8}}^{\ast}$\\ \hline
DF-SVM & $\mathbf{95.01\%\pm0.32\%}$ &  ----\\ \hline
DF-MLP & 93.61$\%$ $\pm$0.43$\%$ & ${6.3\times10^{-8}}^{\ast}$\\ \hline
\end{tabular}
\begin{tablenotes}
\item[1] The bold value represents the highest accuracy in each column.
\item[2] p: The p-value of the Wilcoxon rank-sum test between the performance of DF-SVM and other algorithms.
\item[3] $^{\ast} p<0.05$ 
\end{tablenotes}
\small
\end{table}

\begin{table}[!t]
\centering
\caption{Experiment Result of London Weather Dataset.}
\label{table_london}
\renewcommand{\arraystretch}{1.5}
\begin{tabular}{p{2.5cm}<{\centering}p{2.5cm}<{\centering}p{2.5cm}<{\centering}}
\hline
Algorithms & Test accuracy & p\\ \hline
Meanlogistic  & 71.58$\%$ $\pm$1.94$\%$  &  ${0.0038}^{\ast}$\\ \hline
MeanSVM &  72.26$\%$ $\pm$2.15$\%$   & $0.049^{\ast}$ \\ \hline
MeanDecisiontree & 69.11$\%$ $\pm$1.99$\%$ & ${1.5\times10^{-5}}^{\ast}$ \\ \hline
MeanRandomForest & 72.76$\%$ $\pm$1.84$\%$ & $0.24$ \\ \hline
MeanMLP & 71.53$\%$ $\pm$2.10$\%$ & $0.00059^{\ast}$\\ \hline
DF-SVM & $\mathbf{73.55\%\pm1.73\%}$ &  ----\\ \hline
DF-MLP & 73.06$\%$ $\pm$1.91$\%$ & $0.33$\\ \hline
\end{tabular}
\begin{tablenotes}
\item[1] The bold value represents the highest accuracy in each column.
\item[2] p: The p-value of the Wilcoxon rank-sum test between the performance of DF-SVM and other algorithms.
\item[3] $^{\ast} p<0.05$ 
\end{tablenotes}
\small
\end{table}

\begin{table}[!t]
\centering
\caption{Experiment Result of Washington Weather Dataset.}
\label{table_wash}
\renewcommand{\arraystretch}{1.5}
\begin{tabular}{p{2.5cm}<{\centering}p{2.5cm}<{\centering}p{2.5cm}<{\centering}}
\hline
Algorithms & Test accuracy & p\\ \hline
Meanlogistic  & 97.60$\%$ $\pm$0.60$\%$  &  ${0.045}^{\ast}$\\ \hline
MeanSVM &  97.76$\%$ $\pm$0.66$\%$   & $0.30$ \\ \hline
MeanDecisiontree & 97.26$\%$ $\pm$0.74$\%$ & ${0.0026}^{\ast}$ \\ \hline
MeanRandomForest & 97.34$\%$ $\pm$0.74$\%$ & $0.0043^{\ast}$ \\ \hline
MeanMLP & 97.65$\%$ $\pm$0.52$\%$ & $0.049^{\ast}$\\ \hline
DF-SVM & 97.95$\%$ $\pm$0.66$\%$ & $0.90$\\ \hline
DF-MLP & $\mathbf{98.01\%\pm0.62\%}$ &  ----\\ \hline
\end{tabular}
\begin{tablenotes}
\item[1] The bold value represents the highest accuracy in each column.
\item[2] p: The p-value of the Wilcoxon rank-sum test between the performance of DF-SVM and other algorithms.
\item[3] $^{\ast} p<0.05$ 
\end{tablenotes}
\small
\end{table}

\begin{figure*}[tp]
    \begin{center}
        \subfigure[DF-MLP on the perceptions experiment dataset.]
        {\includegraphics[width=0.32\textwidth]{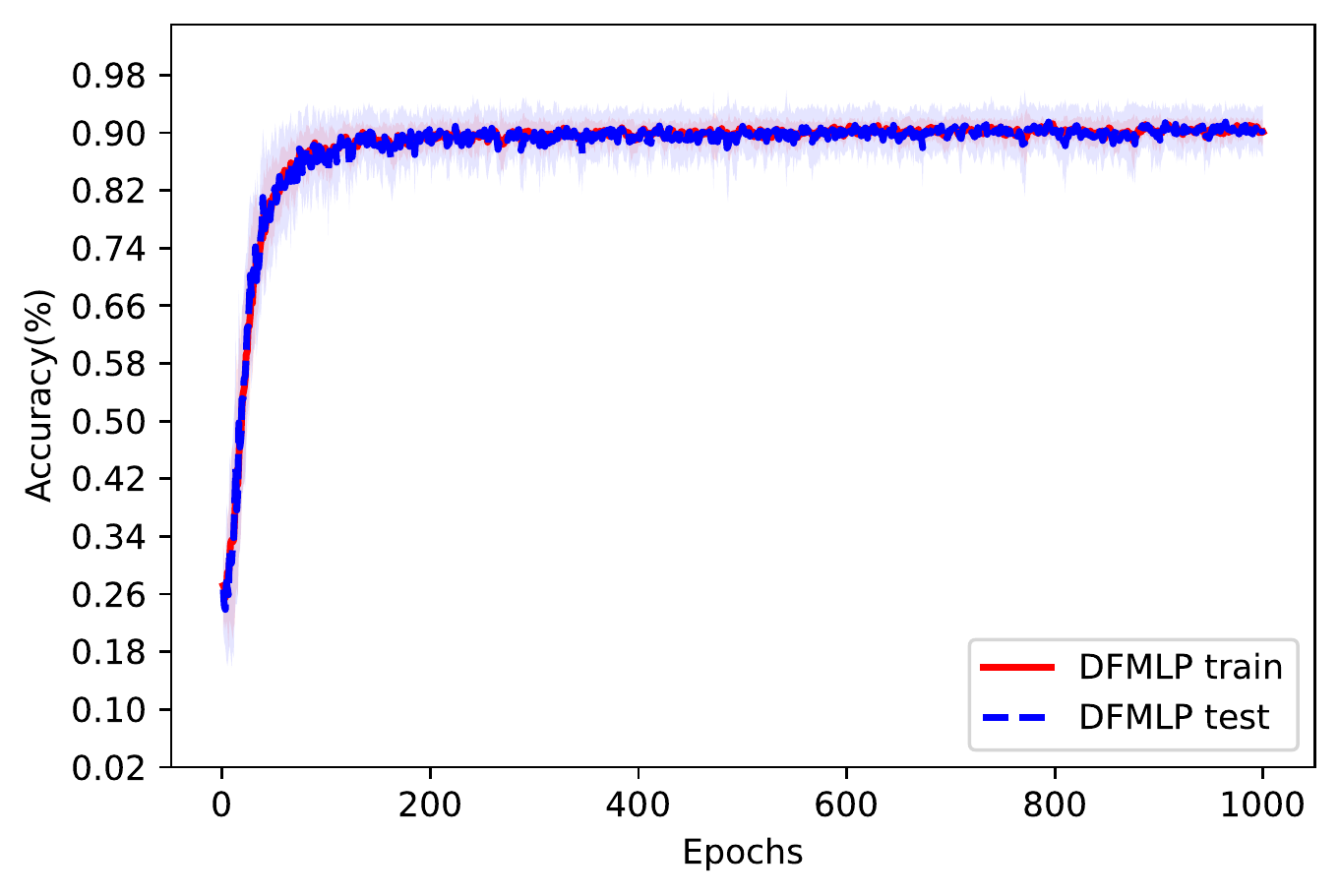}}
        \subfigure[DF-MLP on the mushroom dataset.]
        {\includegraphics[width=0.32\textwidth]{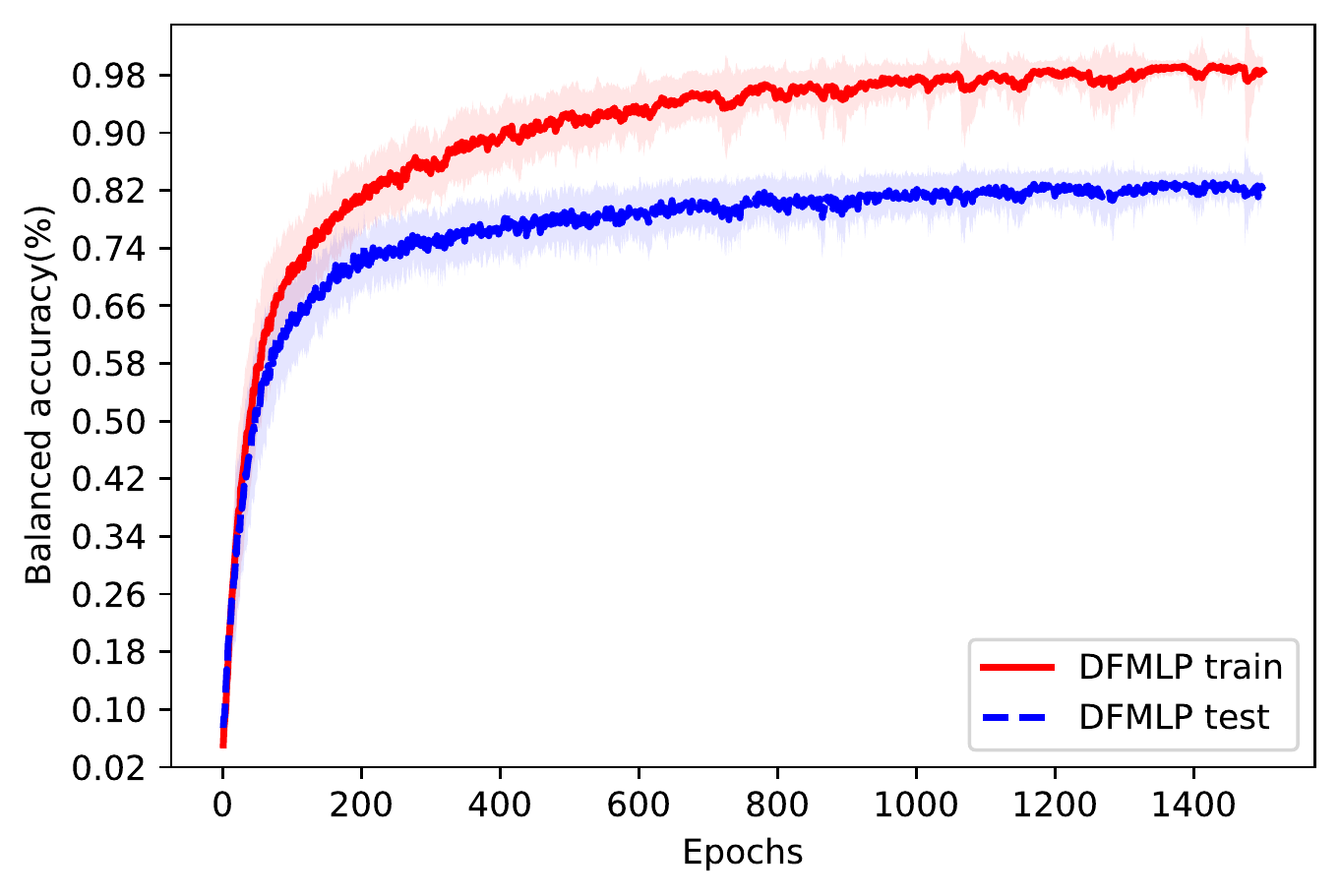}}
        \subfigure[DF-MLP on the mushroom dataset.]
        {\includegraphics[width=0.32\textwidth]{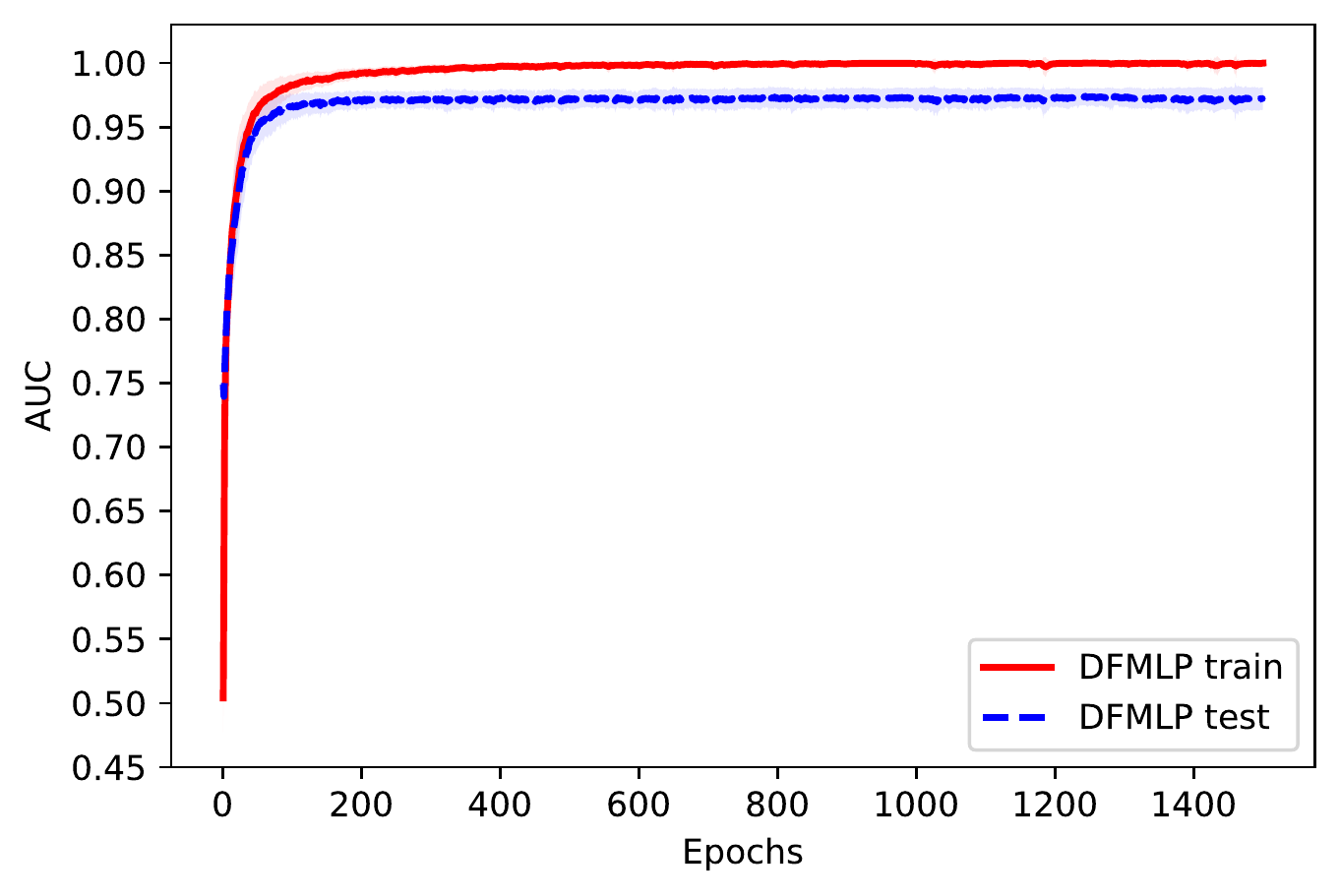}}
        \subfigure[DF-MLP on the letter recognition dataset.]
        {\includegraphics[width=0.32\textwidth]{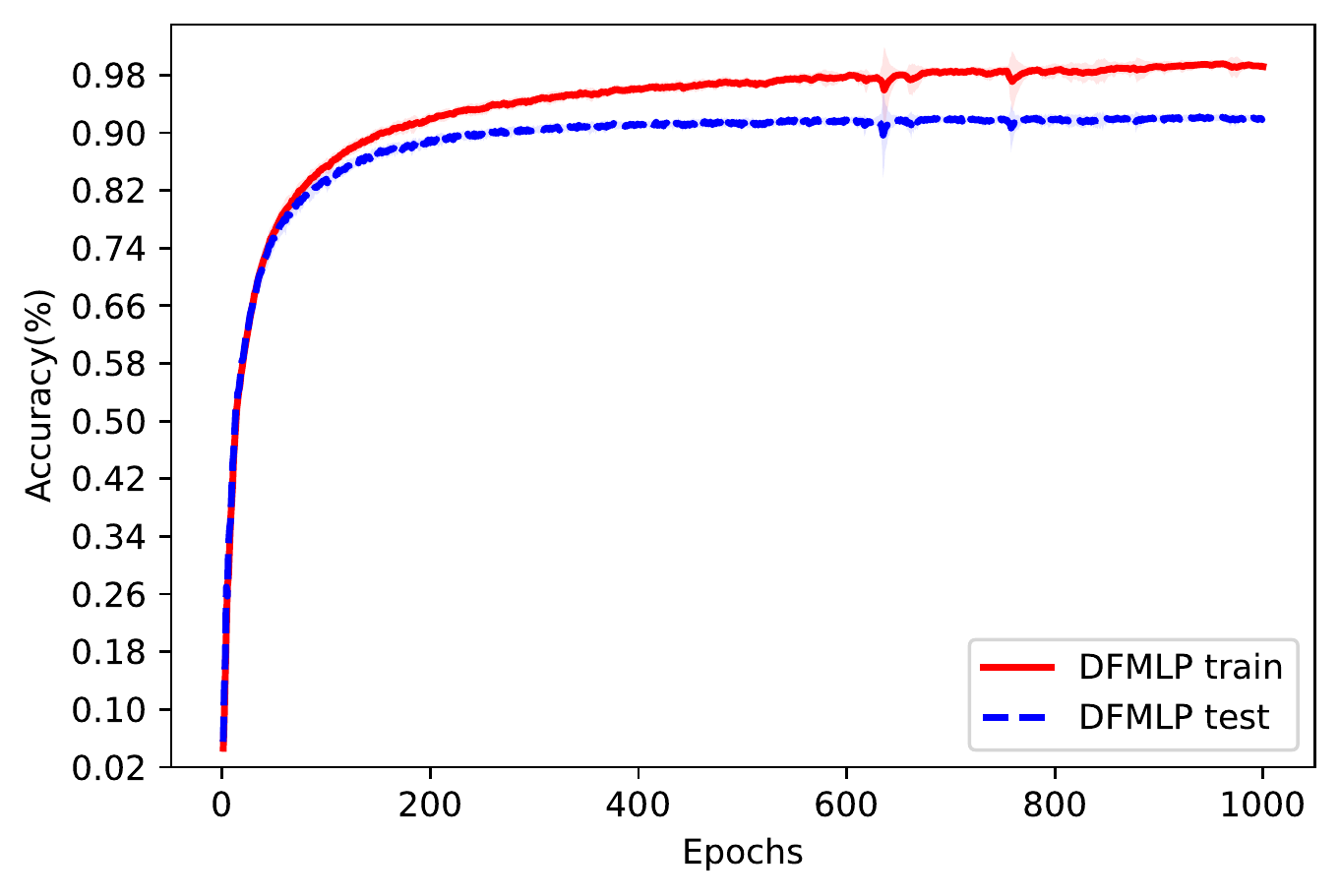}}
        \subfigure[DF-MLP on the London weather dataset.]
        {\includegraphics[width=0.32\textwidth]{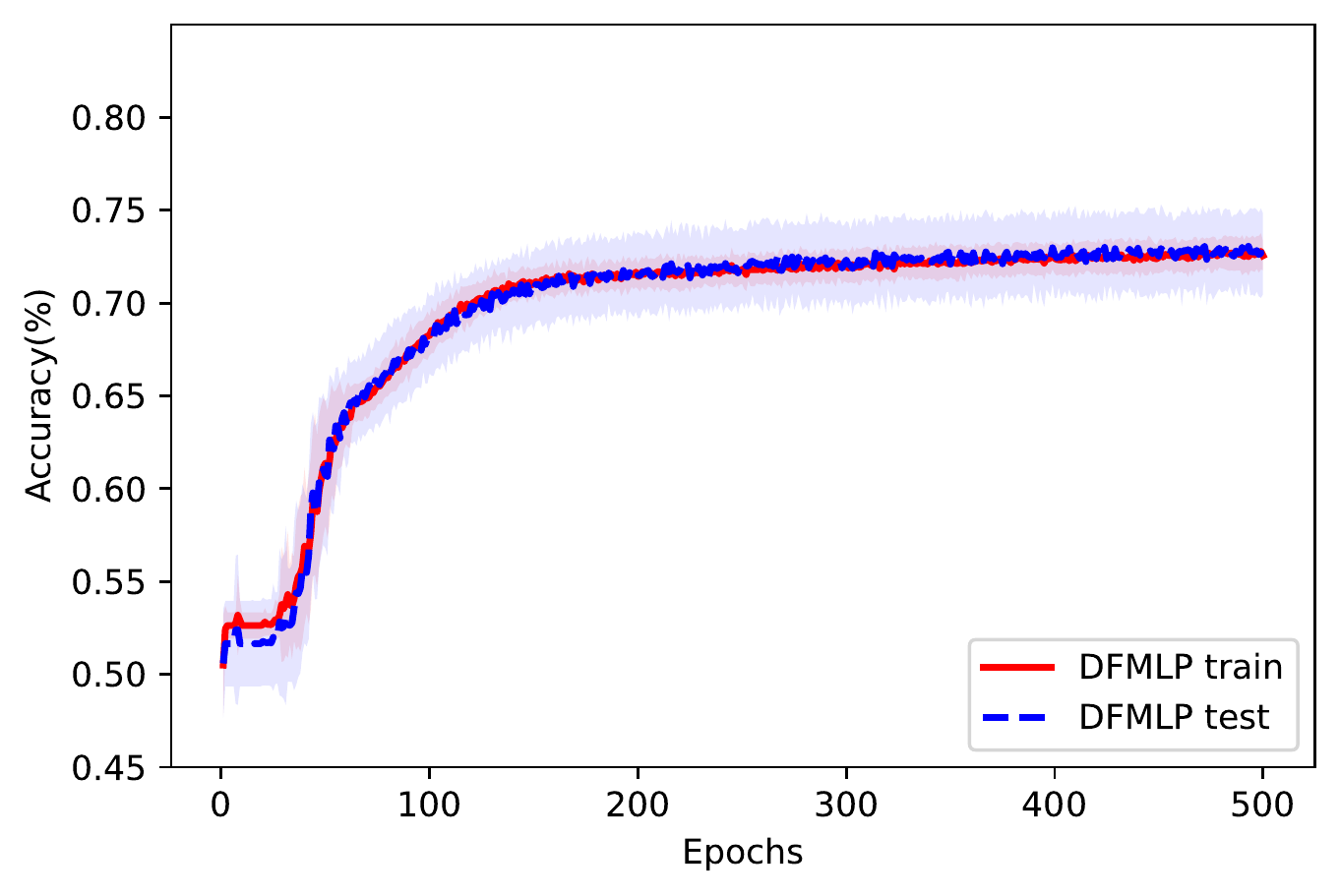}}
        \subfigure[DF-MLP on the Washington weather dataset.]
        {\includegraphics[width=0.32\textwidth]{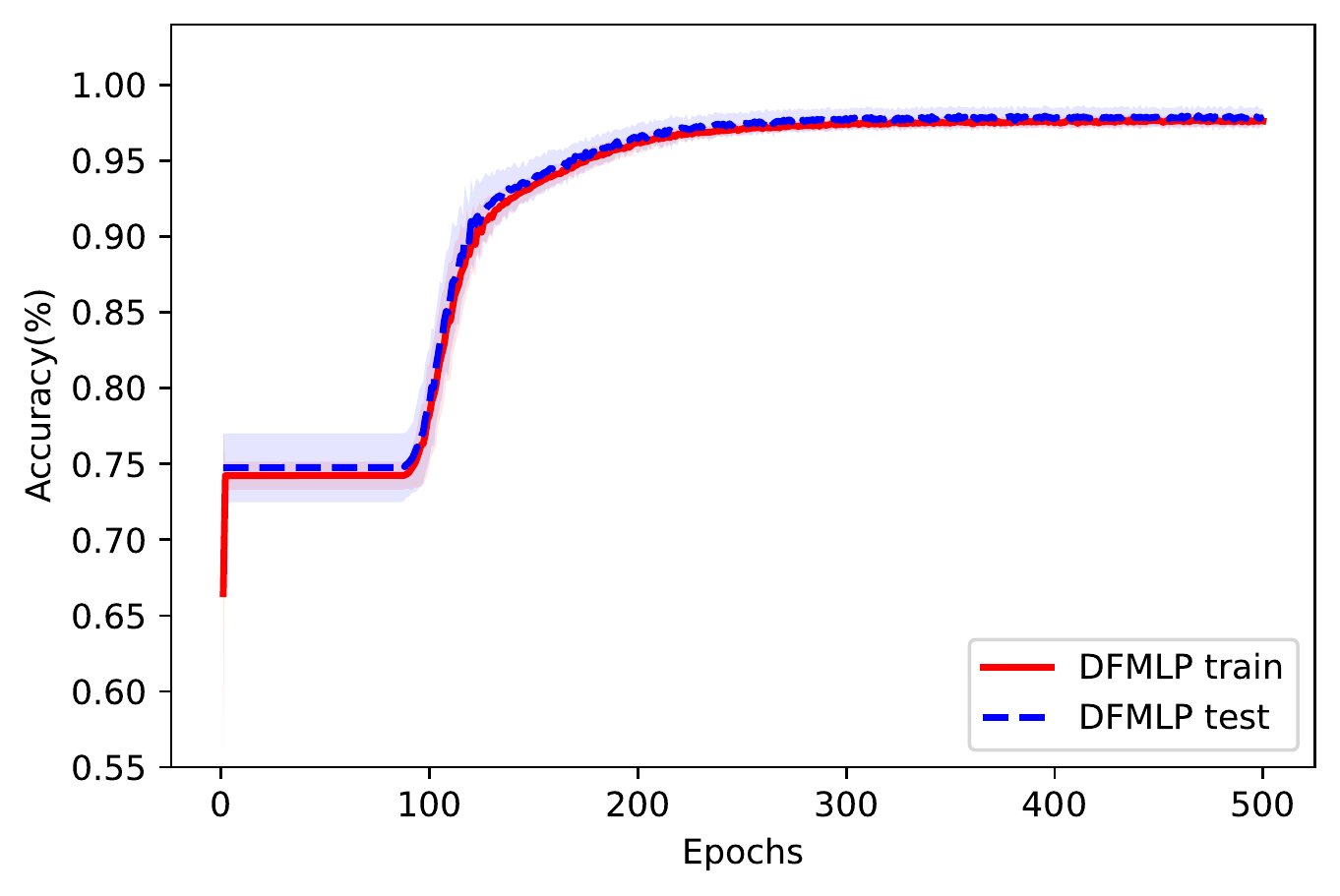}}
        \caption{Evaluation metrics varies with the number of epochs.
        }\label{fig01}
    \end{center}
\end{figure*}

\begin{figure*}[tp]
    \begin{center}
        \subfigure[DF-SVM.]
        {\includegraphics[width=0.32\textwidth]{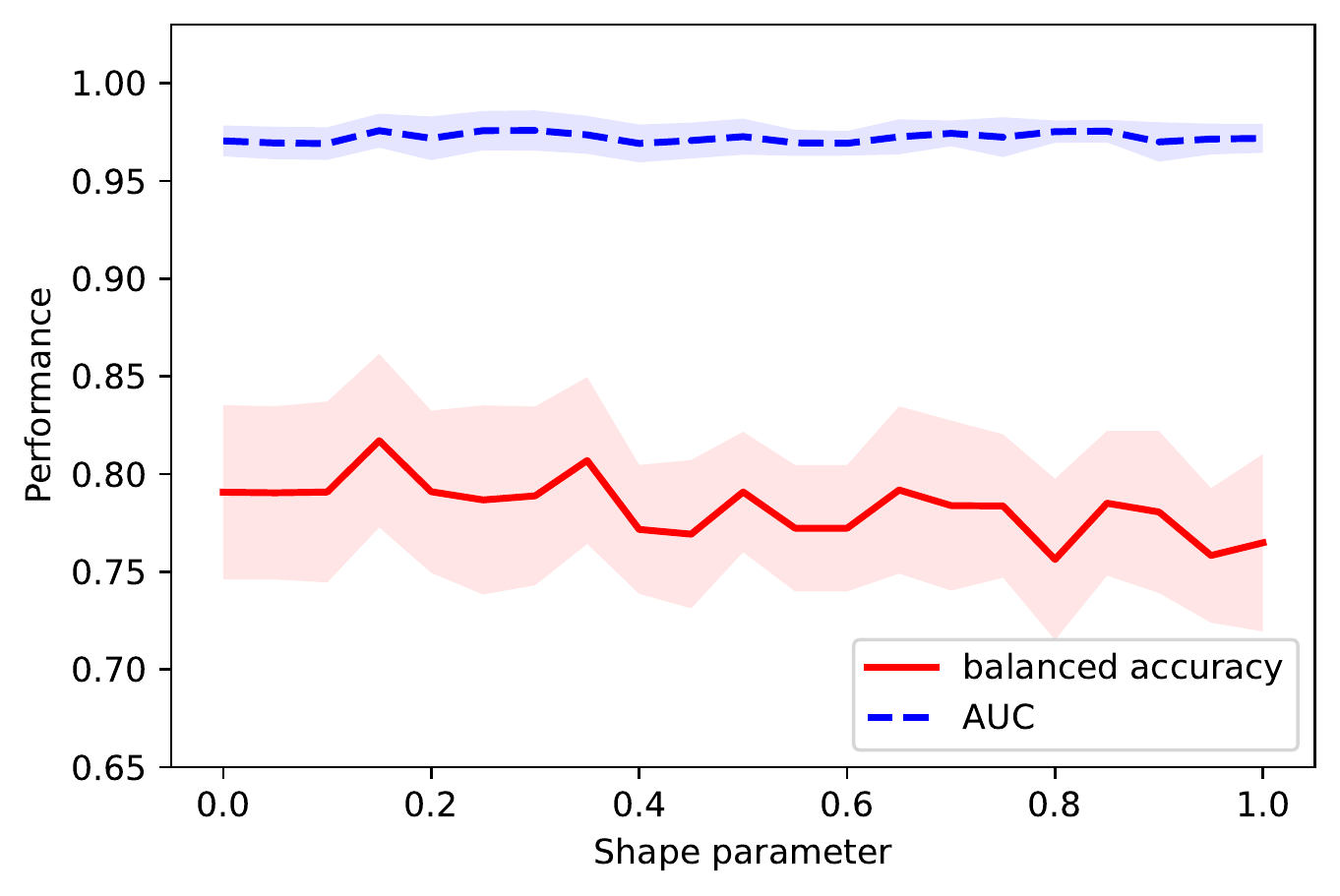}\label{fig_1_case}}
        \subfigure[DF-MLP.]
        {\includegraphics[width=0.32\textwidth]{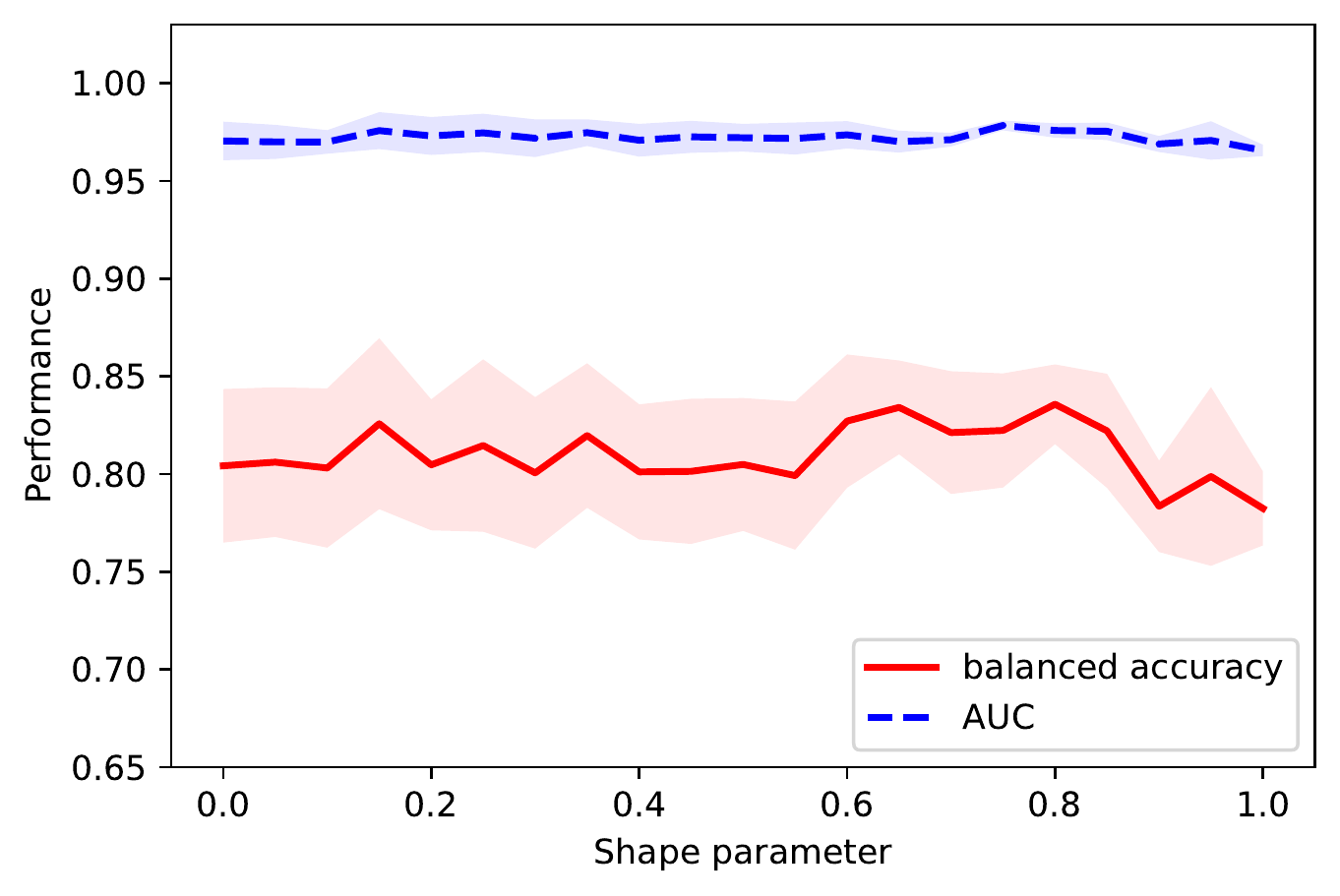}\label{fig_2_case}}
        \caption{Evaluation metrics of the test sets varies with the value of shape parameter $\beta$.
        }\label{fig_sim}
    \end{center}
\end{figure*}

\subsection{Experimental results analysis}
All the experiment results on the five real-world datasets are illustrated in Tables \ref{table_2}, \ref{table_4}, \ref{table_test}, \ref{table_letter}, \ref{table_london}, \ref{table_wash} and how the evaluation metrics varies with the number of epochs for Algorithm \ref{A2} are shown in Figure \ref{fig01}. From these results, the proposed two algorithms achieve better performance than other baselines on all five real-world datasets, which illustrates the efficacy of the proposed algorithms in addressing real-world datasets with fuzzy-valued or interval-valued features. Moreover, the results of the statistic test show that the proposed two algorithms outperform most other methods significantly at the $0.05$ significance level, which demonstrates the superiority of the proposed algorithms. Further, for the $1$st, $2$nd and $5$th datasets, DF-MLP obtains the highest average performance on the test set. While, for the letter recognition dataset and London weather dataset, DF-SVM is more prioritized than other methods, which means that the proposed algorithms are applicable to different types of datasets. 

\subsection{Parameters sensitivity analysis}
In this section, we analyze whether the value of the shape parameter $\beta$ in DF-SVM and DF-MLP affects the balanced accuracy and $AUC$ on the mushroom dataset.

We conduct the same preprocessing for the mushroom dataset. We select the shape parameter $\beta$ from $\{0, 0.05, 0.1, \cdots ,1\}$. Then, for each value of $\beta$, the results are obtained using the same experimental operation in Section \ref{sec:experiment}. Figures \ref{fig_1_case} and \ref{fig_2_case} show the mean and standard deviation of the balanced accuracy and $AUC$ of the test sets on the mushroom dataset when the shape parameter $\beta$ of both algorithms changes from $0$ to $1$. These figures illustrate that a different value for the shape parameter $\beta$ will affect the classification performance since the value of $\beta$ determines the shape of the triangular fuzzy number. A value of $\beta$ that can achieve high performance means that the proposed algorithms with this value of $\beta$ can extract more significant information from the datasets with fuzzy-valued or interval-valued features. Therefore, we can improve the performance of DF-SVM and DF-MLP by finding a suitable value of $\beta$. In our experiments, we find the optimal value of $\beta$ in the validation set.

\section{Conclusion and Future Work}
\label{sec:conclusion}
In this paper, we identify a new problem called \emph{multi-class classification with imprecise observations} (MCIMO). In the MCIMO problem, we need to train a fuzzy classifier when only fuzzy-feature observations are available.

Firstly, we identify a novel problem called MCIMO in Section \ref{sec:MCIMO}. Since there are no existing papers for theoretical analysis of fuzzy classifiers, we give the estimation error bounds for the MCIMO problem in this paper. These bounds illustrate that we can always train a fuzzy classifier with high classification accuracy to solve the MCIMO problem as long as sufficient fuzzy-feature instances can be collected.

Hence, two algorithms are constructed to handle the MCIMO problem. In addition, the optimal defuzzification function for the proposed fuzzy technique-based algorithms is found by comparing the performance of different defuzzification methods on synthetic datasets. Finally, experimental results on synthetic datasets and three real-world datasets show the superiority of the proposed algorithms. Moreover, through comparisons with several non-fuzzy baselines, the experimental results demonstrate that the proposed fuzzy-based methods can obtain better performance in analyzing fuzzy data or interval-valued data than non-fuzzy methods. Since they use fuzzy vectors to express the distribution of imprecise data and apply different defuzzification methods to extract crisp-valued information from imprecise observations.

In future research, we plan to study more complicated issues, for example, covariate shift and domain adaptation with imprecise observations. We can get the theoretical analysis and solutions of these issues based on the introduced theoretical analysis and algorithms in this paper. In addition, we found that the proposed two algorithms can obtain better performance in processing interval-valued data. Therefore, we consider analyzing interval-valued data based on the proposed two algorithms in future studies.

\ifCLASSOPTIONcaptionsoff
  \newpage
\fi



%

\bibliographystyle{IEEEtran}
\bibliography{IEEEabrv,cas-refs}

\begin{thebibliography}{10}
\providecommand{\url}[1]{#1}
\csname url@samestyle\endcsname
\providecommand{\newblock}{\relax}
\providecommand{\bibinfo}[2]{#2}
\providecommand{\BIBentrySTDinterwordspacing}{\spaceskip=0pt\relax}
\providecommand{\BIBentryALTinterwordstretchfactor}{4}
\providecommand{\BIBentryALTinterwordspacing}{\spaceskip=\fontdimen2\font plus
\BIBentryALTinterwordstretchfactor\fontdimen3\font minus
  \fontdimen4\font\relax}
\providecommand{\BIBforeignlanguage}[2]{{%
\expandafter\ifx\csname l@#1\endcsname\relax
\typeout{** WARNING: IEEEtran.bst: No hyphenation pattern has been}%
\typeout{** loaded for the language `#1'. Using the pattern for}%
\typeout{** the default language instead.}%
\else
\language=\csname l@#1\endcsname
\fi
#2}}
\providecommand{\BIBdecl}{\relax}
\BIBdecl

\bibitem{2019seebock}
P.~{Seeböck}, S.~M. {Waldstein}, S.~{Klimscha}, H.~{Bogunovic}, T.~{Schlegl},
  B.~S. {Gerendas}, R.~{Donner}, U.~{Schmidt-Erfurth}, and G.~{Langs},
  ``Unsupervised identification of disease marker candidates in retinal {OCT}
  imaging data,'' \emph{IEEE Transactions on Medical Imaging}, vol.~38, no.~4,
  pp. 1037--1047, 2019.

\bibitem{2013xia}
R.~{Xia}, C.~{Zong}, X.~{Hu}, and E.~{Cambria}, ``Feature ensemble plus sample
  selection: Domain adaptation for sentiment classification,'' \emph{IEEE
  Intelligent Systems}, vol.~28, no.~3, pp. 10--18, 2013.

\bibitem{2016zhu}
X.~{Zhu}, H.~{Suk}, S.~{Lee}, and D.~{Shen}, ``Subspace regularized sparse
  multitask learning for multiclass neurodegenerative disease identification,''
  \emph{IEEE Transactions on Biomedical Engineering}, vol.~63, no.~3, pp.
  607--618, 2016.

\bibitem{nanni2017handcrafted}
L.~Nanni, S.~Ghidoni, and S.~Brahnam, ``Handcrafted vs. non-handcrafted
  features for computer vision classification,'' \emph{Pattern Recognition},
  vol.~71, pp. 158--172, 2017.

\bibitem{wang2018transfer}
G.~Wang, J.~Lu, K.-S. Choi, and G.~Zhang, ``A transfer-based additive ls-svm
  classifier for handling missing data,'' \emph{IEEE transactions on
  cybernetics}, vol.~50, no.~2, pp. 739--752, 2018.

\bibitem{chen2013fuzzy}
C.~P. Chen, Y.-J. Liu, and G.-X. Wen, ``Fuzzy neural network-based adaptive
  control for a class of uncertain nonlinear stochastic systems,'' \emph{IEEE
  Transactions on Cybernetics}, vol.~44, no.~5, pp. 583--593, 2013.

\bibitem{Mehryar2012}
M.~Mohri, A.~Rostamizadeh, and A.~Talwalkar, \emph{Foundations of Machine
  Learning}.\hskip 1em plus 0.5em minus 0.4em\relax Massachusetts Institute of
  Technology, 2012.

\bibitem{Panchenko2002Empirical}
V.~Koltchinskii and D.~Panchenko, ``Empirical margin distributions and bounding
  the generalization error of combined classifiers,'' \emph{Annals of
  Statistics}, vol.~30, no.~1, pp. 1--50, 2002.

\bibitem{Maximov2018}
Y.~Maximov, M.-R. Amini, and Z.~Harchaoui, ``Rademacher complexity bounds for a
  penalized multi-class semi-supervised algorithm,'' \emph{Journal of
  Artificial Intelligence Research}, vol.~61, no.~1, p. 761–786, Jan. 2018.

\bibitem{Allwein2000Reducing}
E.~L. Allwein, R.~E. Schapire, and Y.~Singer, ``Reducing multiclass to binary:
  A unifying approach for margin classifiers,'' \emph{Journal of Machine
  Learning Research}, vol.~1, no.~2, 2000.

\bibitem{2014Optimal}
A.~Daniely and S.~Shalev-Shwartz, ``Optimal learners for multiclass problems,''
  in \emph{Conference on Learning Theory}, 2014, pp. 287--316.

\bibitem{hardt2016train}
M.~Hardt, B.~Recht, and Y.~Singer, ``Train faster, generalize better: Stability
  of stochastic gradient descent,'' in \emph{International Conference on
  Machine Learning}, 2016, pp. 1225--1234.

\bibitem{mcallester2013pac}
D.~McAllester, ``A {PAC}-bayesian tutorial with a dropout bound,''
  \emph{Computer Science}, 2013.

\bibitem{Xu2016}
C.~{Xu}, T.~{Liu}, D.~{Tao}, and C.~{Xu}, ``Local rademacher complexity for
  multi-label learning,'' \emph{IEEE Transactions on Image Processing},
  vol.~25, no.~3, pp. 1495--1507, 2016.

\bibitem{Li2018}
J.~Li, Y.~Liu, R.~Yin, H.~Zhang, L.~Ding, and W.~Wang, ``Multi-class learning:
  From theory to algorithm.'' \emph{NeurIPS}, vol.~31, pp. 1593--1602, 2018.

\bibitem{zhang2020clarinet}
Y.~Zhang, F.~Liu, Z.~Fang, B.~Yuan, G.~Zhang, and J.~Lu, ``Clarinet: A one-step
  approach towards budget-friendly unsupervised domain adaptation,''
  \emph{arXiv preprint arXiv:2007.14612}, 2020.

\bibitem{fang2020open}
Z.~Fang, J.~Lu, F.~Liu, J.~Xuan, and G.~Zhang, ``Open set domain adaptation:
  Theoretical bound and algorithm,'' \emph{IEEE Transactions on Neural Networks
  and Learning Systems}, 2020.

\bibitem{zhong2021does}
L.~Zhong, Z.~Fang, F.~Liu, J.~Lu, B.~Yuan, and G.~Zhang, ``How does the
  combined risk affect the performance of unsupervised domain adaptation
  approaches?'' in \emph{Proceedings of the 35th AAAI Conference on Artificial
  Intelligence}, 2021.

\bibitem{zhen2022heterogeneous}
Z.~Fang, J.~Lu, F.~Liu, and G.~Zhang, ``Semi-supervised heterogeneous domain
  adaptation: Theory and algorithms,'' \emph{IEEE Transactions on Pattern
  Analysis and Machine Intelligence}, 2022.

\bibitem{jiahua2022domain}
J.~Dong, Y.~Cong, G.~Sun, Z.~Fang, and Z.~Ding, ``Where and how to transfer:
  Knowledge aggregation-induced transferability perception for unsupervised
  domain adaptation,'' \emph{IEEE Transactions on Pattern Analysis and Machine
  Intelligence}, 2021.

\bibitem{Colubi2011Nonparametric}
A.~Colubi, G.~González-Rodríguez, M.~Ángeles Gil, and W.~Trutschnig,
  ``Nonparametric criteria for supervised classification of fuzzy data,''
  \emph{International Journal of Approximate Reasoning}, vol.~52, no.~9, pp.
  1272--1282, 2011.

\bibitem{wang2020deep}
G.~Wang, T.~Zhou, K.-S. Choi, and J.~Lu, ``A deep-ensemble-level-based
  interpretable takagi-sugeno-kang fuzzy classifier for imbalanced data,''
  \emph{IEEE transactions on cybernetics}, 2020.

\bibitem{liu2020multi}
F.~Liu, G.~Zhang, and J.~Lu, ``Multi-source heterogeneous unsupervised domain
  adaptation via fuzzy-relation neural networks,'' \emph{IEEE Transactions on
  Fuzzy Systems}, 2020.

\bibitem{zuo2019fuzzy2}
H.~Zuo, J.~Lu, G.~Zhang, and F.~Liu, ``Fuzzy transfer learning using an
  infinite gaussian mixture model and active learning,'' \emph{IEEE
  Transactions on Fuzzy Systems}, vol.~27, no.~2, pp. 291--303, 2019.

\bibitem{lu2019fuzzy}
J.~Lu, H.~Zuo, and G.~Zhang, ``Fuzzy multiple-source transfer learning,''
  \emph{IEEE Transactions on Fuzzy Systems}, vol.~28, no.~12, pp. 3418--3431,
  2020.

\bibitem{ma2021learning}
G.~Ma, F.~Liu, G.~Zhang, and J.~Lu, ``Learning from imprecise observations: An
  estimation error bound based on fuzzy random variables,'' in \emph{2021 IEEE
  International Conference on Fuzzy Systems}, 2021, pp. 1--8.

\bibitem{Ralescu1986}
M.~L. Puri and D.~A. Ralescu, ``Fuzzy random variables,'' \emph{Journal of
  Mathematical Analysis Applications}, vol. 114, no.~2, pp. 409--422, 1986.

\bibitem{Wu1999Probability}
H.~C. Wu, ``Probability density functions of fuzzy random variables,''
  \emph{Fuzzy Sets and Systems}, vol. 105, no.~1, pp. 139--158, 1999.

\bibitem{Sinova2014A}
B.~Sinova, M.~Ángeles Gil, M.~T. López, and S.~V. Aelst, ``A parameterized
  {$L_2$} metric between fuzzy numbers and its parameter interpretation,''
  \emph{Fuzzy Sets and Systems}, vol. 245, pp. 101--115, 2014.

\bibitem{Yang2011fuzzysvm}
X.~Yang, G.~Zhang, J.~Lu, and J.~Ma, ``A kernel fuzzy {C}-means
  clustering-based fuzzy support vector machine algorithm for classification
  problems with outliers or noises,'' \emph{IEEE Transactions on Fuzzy
  Systems}, vol.~19, no.~1, pp. 105--115, 2011.

\bibitem{Rong2007Classification}
Y.~Rong, Z.~Wang, P.~A. Heng, and K.~S. Leung, ``Classification of
  heterogeneous fuzzy data by choquet integral with fuzzy-valued integrand,''
  \emph{IEEE Transactions on Fuzzy Systems}, vol.~15, no.~5, pp. 931--942,
  2007.

\bibitem{liu2020anovel}
F.~Liu, G.~Zhang, and J.~Lu, ``A novel non-parametric two-sample test on
  imprecise observations,'' in \emph{FUZZ-IEEE}, 2020, pp. 1--6.

\bibitem{gretton2012kernel}
A.~Gretton, K.~M. Borgwardt, M.~J. Rasch, B.~Sch{\"o}lkopf, and A.~Smola, ``A
  kernel two-sample test,'' \emph{The Journal of Machine Learning Research},
  vol.~13, no.~1, pp. 723--773, 2012.

\bibitem{liu2020learning}
F.~Liu, W.~Xu, J.~Lu, G.~Zhang, A.~Gretton, and D.~J. Sutherland, ``Learning
  deep kernels for non-parametric two-sample tests,'' in \emph{ICML}, 2020, pp.
  6316--6326.

\bibitem{liu2021meta}
F.~Liu, W.~Xu, J.~Lu, and D.~J. Sutherland, ``Meta two-sample testing: Learning
  kernels for testing with limited data,'' in \emph{NeurIPS}, 2021, pp.
  5848--5860.

\bibitem{behbood2014fuzzy}
{V. Behbood, J. Lu, and G. Zhang,}, ``Fuzzy refinement domain adaptation for
  long term prediction in banking ecosystem,'' \emph{IEEE Transactions on
  Industrial Informatics}, vol.~10, no.~2, pp. 1637--1646, 2014.

\bibitem{behbood2015multistep}
V.~Behbood, J.~Lu, G.~Zhang, and W.~Pedrycz, ``Multistep fuzzy bridged
  refinement domain adaptation algorithm and its application to bank failure
  prediction,'' \emph{IEEE Transactions on Fuzzy Systems}, vol.~23, no.~6, pp.
  1917--1935, 2015.

\bibitem{yang2015takagi}
C.~Yang, Z.~Deng, K.-S. Choi, and S.~Wang, ``{Takagi--Sugeno--Kang} transfer
  learning fuzzy logic system for the adaptive recognition of epileptic
  electroencephalogram signals,'' \emph{IEEE Transactions on Fuzzy Systems},
  vol.~24, no.~5, pp. 1079--1094, 2016.

\bibitem{deng2018transductive}
Z.~Deng, P.~Xu, L.~Xie, K.-S. Choi, and S.~Wang, ``Transductive
  joint-knowledge-transfer {TSK FS} for recognition of epileptic {EEG}
  signals,'' \emph{IEEE Transactions on Neural Systems and Rehabilitation
  Engineering}, vol.~26, no.~8, pp. 1481--1494, 2018.

\bibitem{xie2018generalized}
L.~Xie, Z.~Deng, P.~Xu, K.-S. Choi, and S.~Wang, ``Generalized hidden-mapping
  transductive transfer learning for recognition of epileptic
  electroencephalogram signals,'' \emph{IEEE Transactions on Cybernetics},
  vol.~49, no.~6, pp. 2200--2214, 2018.

\bibitem{jiang2021eeg}
Y.~Jiang, Y.~Zhang, C.~Lin, D.~Wu, and C.-T. Lin, ``{EEG}-based driver
  drowsiness estimation using an online multi-view and transfer {TSK} fuzzy
  system,'' \emph{IEEE Transactions on Intelligent Transportation Systems},
  2021.

\bibitem{liu2018unsupervised}
F.~Liu, J.~Lu, and G.~Zhang, ``Unsupervised heterogeneous domain adaptation via
  shared fuzzy equivalence relations,'' \emph{IEEE Transactions on Fuzzy
  Systems}, vol.~26, no.~6, pp. 3555--3568, 2018.

\bibitem{deng2012knowledge}
Z.~Deng, Y.~Jiang, F.-L. Chung, H.~Ishibuchi, and S.~Wang,
  ``Knowledge-leverage-based fuzzy system and its modeling,'' \emph{IEEE
  Transactions on Fuzzy Systems}, vol.~21, no.~4, pp. 597--609, 2013.

\bibitem{deng2013knowledge}
Z.~Deng, Y.~Jiang, K.-S. Choi, F.-L. Chung, and S.~Wang,
  ``Knowledge-leverage-based {TSK} fuzzy system modeling,'' \emph{IEEE
  Transactions on Neural Networks and Learning Systems}, vol.~24, no.~8, pp.
  1200--1212, 2013.

\bibitem{deng2016enhanced}
Z.~Deng, Y.~Jiang, H.~Ishibuchi, K.-S. Choi, and S.~Wang, ``Enhanced
  knowledge-leverage-based {TSK} fuzzy system modeling for inductive transfer
  learning,'' \emph{ACM Transactions on Intelligent Systems and Technology},
  vol.~8, no.~1, pp. 1--21, 2016.

\bibitem{zuo2018granular}
H.~Zuo, G.~Zhang, W.~Pedrycz, V.~Behbood, and J.~Lu, ``Granular fuzzy
  regression domain adaptation in {Takagi--Sugeno} fuzzy models,'' \emph{IEEE
  Transactions on Fuzzy Systems}, vol.~26, no.~2, pp. 847--858, 2018.

\bibitem{Benjamin2016Clustering}
B.~Quost and T.~Denœux, ``Clustering and classification of fuzzy data using
  the fuzzy {EM} algorithm,'' \emph{Fuzzy Sets and Systems}, vol. 286, pp.
  134--156, 2016.

\bibitem{Pierpaolo2020Fuzzy}
P.~D'Urso and J.~M. Leski, ``Fuzzy clustering of fuzzy data based on robust
  loss functions and ordered weighted averaging,'' \emph{Fuzzy Sets and
  Systems}, vol. 389, pp. 1--28, 2020.

\bibitem{roychowdhury2001survey}
S.~Roychowdhury and W.~Pedrycz, ``A survey of defuzzification strategies,''
  \emph{International Journal of Intelligent Systems}, vol.~16, no.~6, pp.
  679--695, 2001.

\bibitem{van1999defuzzification}
W.~Van~Leekwijck and E.~E. Kerre, ``Defuzzification: criteria and
  classification,'' \emph{Fuzzy Sets and Systems}, vol. 108, no.~2, pp.
  159--178, 1999.

\bibitem{oussalah2002compatibility}
M.~Oussalah, ``On the compatibility between defuzzification and fuzzy
  arithmetic operations,'' \emph{Fuzzy Sets and Systems}, vol. 128, no.~2, pp.
  247--260, 2002.

\bibitem{delgado1998canonical}
M.~Delgado, M.~A. Vila, and W.~Voxman, ``On a canonical representation of fuzzy
  numbers,'' \emph{Fuzzy Sets and Systems}, vol.~93, no.~1, pp. 125--135, 1998.

\bibitem{weston1999support}
J.~Weston, C.~Watkins \emph{et~al.}, ``Support vector machines for multi-class
  pattern recognition.'' in \emph{European Symposium on Artificial Neural
  Networks}, vol.~99, 1999, pp. 219--224.

\bibitem{kingma2015adam}
D.~P. Kingma and J.~Ba, ``Adam: A method for stochastic optimization,'' in
  \emph{International Conference on Learning Representations}, 2015.

\bibitem{last2017oversampling}
F.~Last, G.~Douzas, and F.~Bacao, ``Oversampling for imbalanced learning based
  on k-means and smote,'' \emph{arXiv preprint arXiv:1711.00837}, 2017.

\bibitem{brodersen2010balanced}
K.~H. Brodersen, C.~S. Ong, K.~E. Stephan, and J.~M. Buhmann, ``The balanced
  accuracy and its posterior distribution,'' in \emph{2010 20th international
  conference on pattern recognition}.\hskip 1em plus 0.5em minus 0.4em\relax
  IEEE, 2010, pp. 3121--3124.

\end{thebibliography}

\vspace{-2cm}
\begin{IEEEbiography}[{\includegraphics[width=1in,height=1.25in,clip,keepaspectratio]{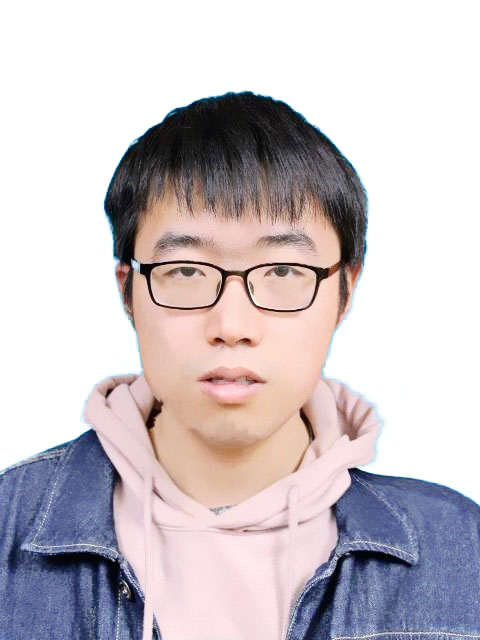}}]{Guangzhi Ma}
received the B.S. degree in mathematics and applied mathematics from the School of Mathematics Sciences, Anhui University, Anhui, China, in 2017 and the M.S. degree in probability and statistics from the School of Mathematics and Statistics, Lanzhou University, Lanzhou, China, in 2020. He is the first year PhD degree with the Faculty of Engineering and Information Technology, University of Technology Sydney, Australia. He is a Member of the Decision Systems and e-Service Intelligence Lab, Australia Artificial Intelligence Institute, University of Technology Sydney. His research interests include fuzzy transfer learning and domain adaptation.
\end{IEEEbiography}

\vspace{-1cm}
\begin{IEEEbiography}[{\includegraphics[width=1in,height=1.25in,clip,keepaspectratio]{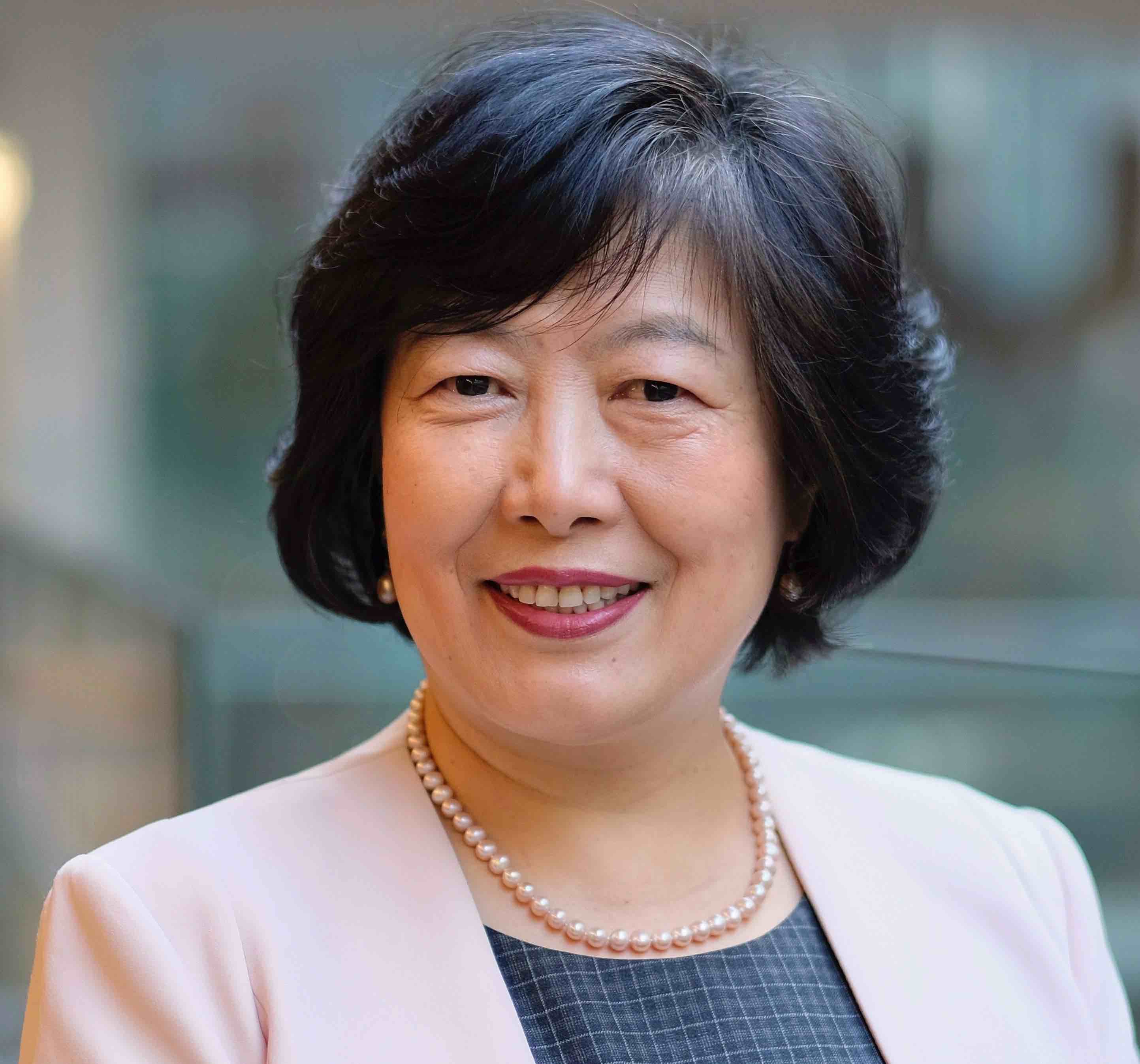}}]{Jie Lu} (F’18) is a Distinguished Professor and the Director of Australian Artificial Intelligence Institute (AAII) at the University of Technology Sydney, Australia. She is also an IFSA Fellow and Australian Laureate Fellow. She received a PhD degree from Curtin University, Australia, in 2000. Her main research expertise is in transfer learning, concept drift, decision support systems and recommender systems. She has been awarded 10+ Australian Research Council (ARC) discovery grants and led 20 industry projects. She has published over 500 papers in IEEE transactions and other journals and conferences, supervised 50 PhD students to completion. She serves as Editor-In-Chief for Knowledge-Based Systems (Elsevier) and Editor-In-Chief for International Journal on Computational Intelligence Systems (Springer). She has delivered 35 keynote speeches at international conferences. She has received the UTS Medal for research excellence (2019), the IEEE Transactions on Fuzzy Systems Outstanding Paper Award (2019) and the Australian Most Innovative Engineer Award (2019).
\end{IEEEbiography}

\vspace{-1cm}
\begin{IEEEbiography}[{\includegraphics[width=1in,height=1.25in,clip,keepaspectratio]{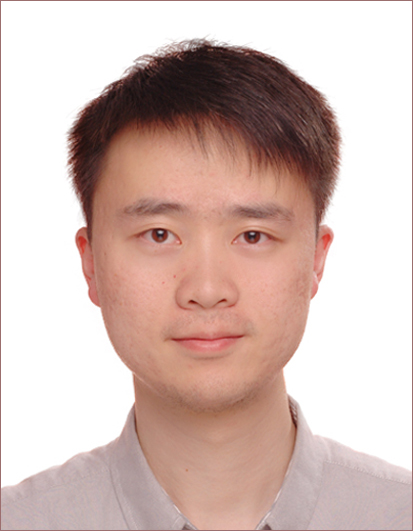}}]{Feng Liu}
is a Lecturer in Australian Artificial Intelligence Institute, Faculty of Engineering and Information Technology, University of Technology Sydney, Australia. He received his Ph.D. degree in computer science from the University of Technology Sydney, and an M.Sc. degree in probability and statistics and a B.Sc. degree in mathematics from the School of Mathematics and Statistics, Lanzhou University, China, in 2015 and 2013, respectively. His research interests include hypothesis testing and trustworthy machine learning. He has served as a senior program committee member for ECAI and program committee members for NeurIPS, ICML, AISTATS, ICLR, KDD, AAAI, IJCAI and FUZZ-IEEE. He also served as reviewers for JMLR, MLJ, TPAMI, TNNLS and TFS. He has received the outstanding reviewer awards of ICLR (2021) and NeurIPS (2021), the UTS-FEIT HDR Research Excellence Award (2019) and Best Student Paper Award of FUZZ-IEEE (2019).
\end{IEEEbiography}

\vspace{-1cm}
\begin{IEEEbiography}[{\includegraphics[width=1.0in,height=1.8in,clip,keepaspectratio]{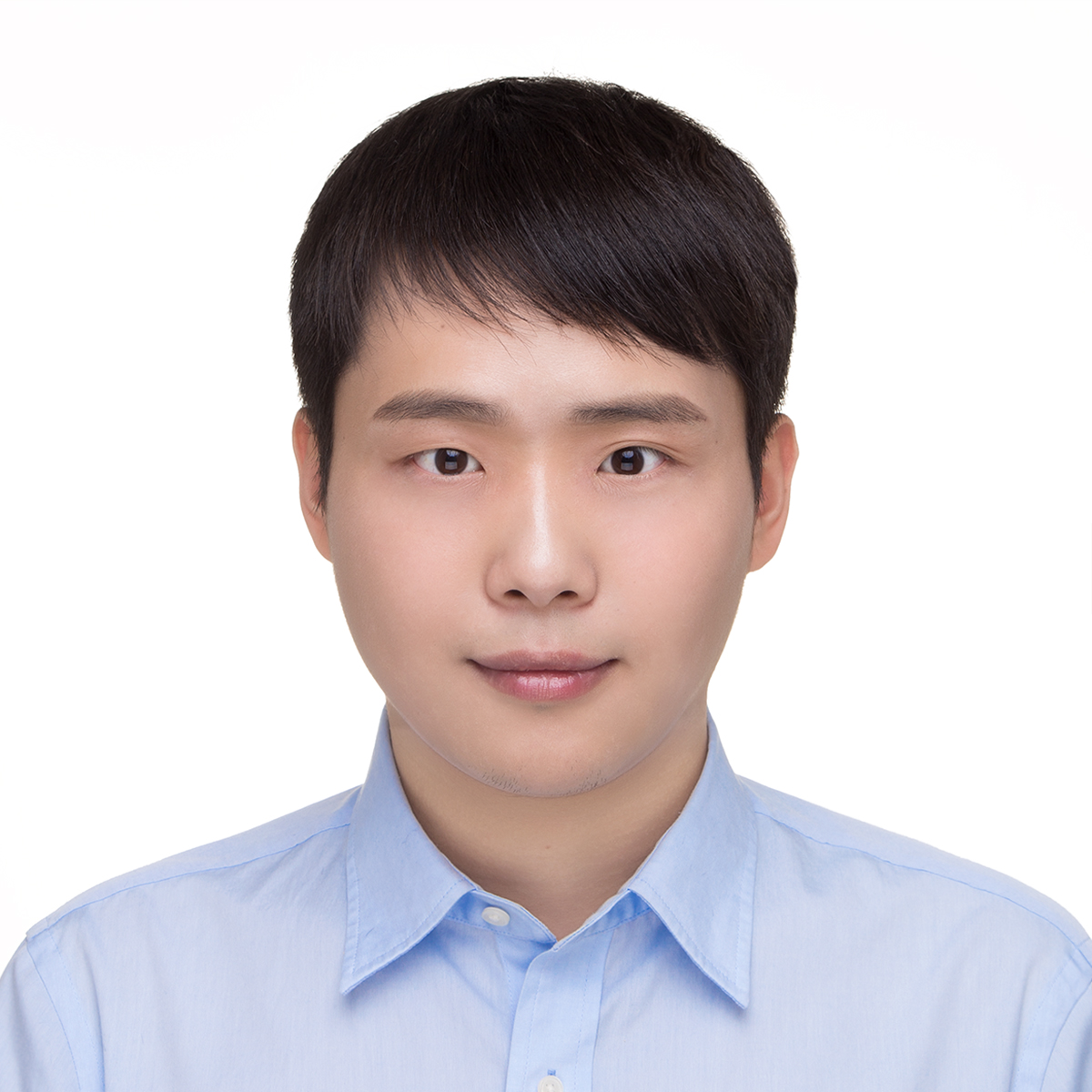}}]{Zhen Fang}
received his M.Sc. degree in pure mathematics from the School of Mathematical Sciences Xiamen University, Xiamen, China, in 2017. He is the final year PhD with the Faculty of Engineering and Information Technology, University of Technology Sydney, Australia. He is a Member of the Decision Systems and e-Service Intelligence (DeSI) Research Laboratory, Australian Artificial Intelligence Institute, University of Technology Sydney. His research interests include transfer learning and out-of-distribution learning. He has published several paper related to transfer learning and out-of-distribution learning in IJCNN, NeurIPS, AAAI, IJCAI, ICML, TNNLS, TPAMI.
\end{IEEEbiography}

\vspace{-1cm}
\begin{IEEEbiography}[{\includegraphics[width=1in,height=1.25in,clip,keepaspectratio]{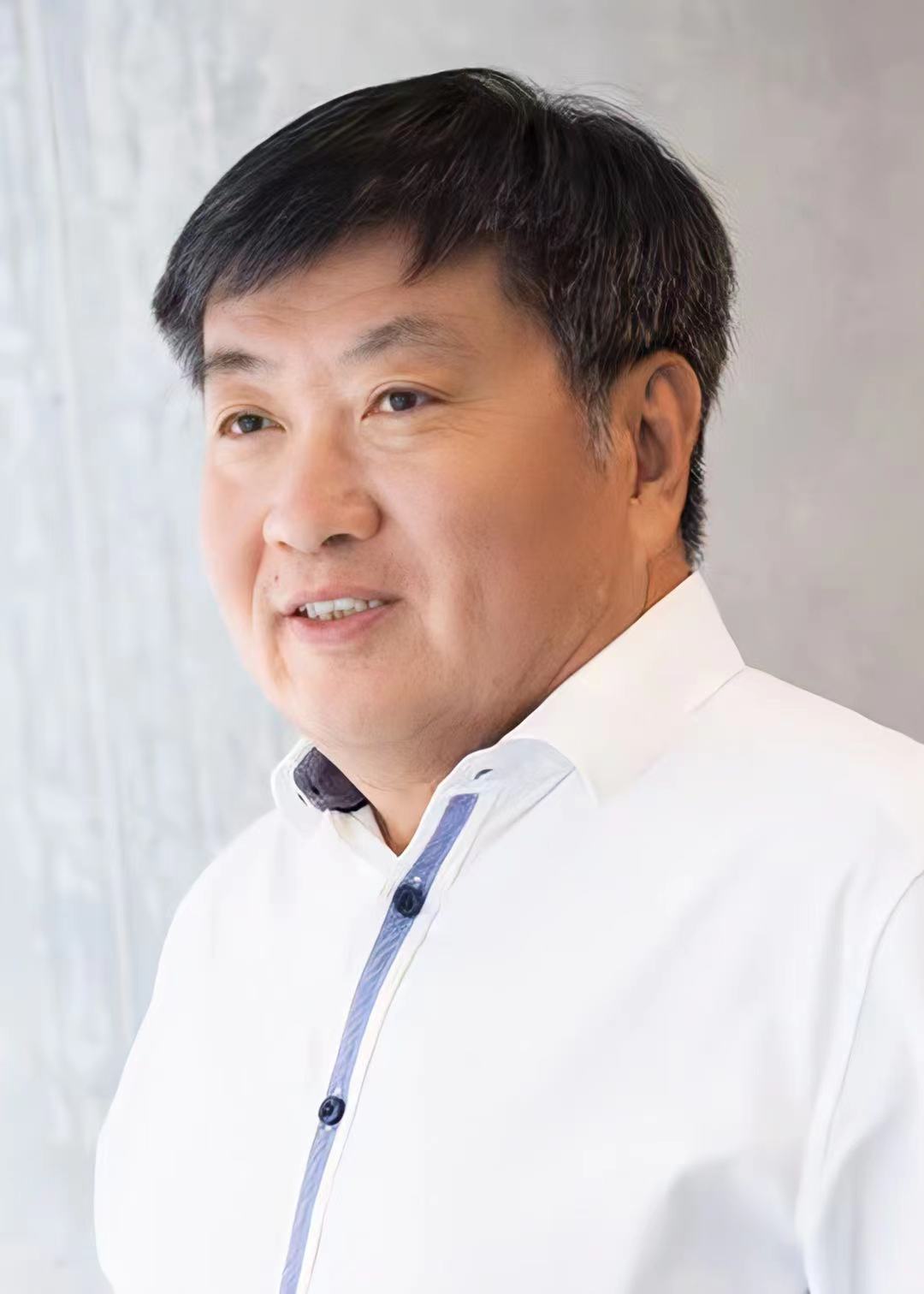}}]{Guangquan Zhang}
is an Australian Research Council (ARC) QEII Fellow, Associate Professor and the Director of the Decision Systems and e-Service Intelligent (DeSI) Research Laboratory at the Australian Artificial Intelligence Institute, University of Technology Sydney, Australia. He received his Ph.D in applied mathematics from Curtin University, Australia, in 2001. From 1993 to 1997, he was a full Professor in the Department of Mathematics, Hebei University, China. His main research interests lie in the area of fuzzy multi-objective, bilevel and group decision making, fuzzy measure, and machine learning. He has published six authored monographs, five edited research books, and over 500 papers including some 300 refereed journal articles. Dr. Zhang has won ten ARC Discovery Project grants and many other research grants, supervised 35 PhD students to completion. He has served as a Guest Editor for special issues of IEEE Transactions and other international journals.
\end{IEEEbiography}

%








\end{document}